\documentclass[10pt,twocolumn,letterpaper]{article}

\usepackage{iccv}
\usepackage{times}
\usepackage{epsfig}
\usepackage{graphicx}
\usepackage{amsmath}
\usepackage{amssymb}

\usepackage{booktabs}
\usepackage{multirow}
\usepackage{graphics}
\usepackage{makecell}
\usepackage{subcaption}
\usepackage{bbm}

\usepackage{algorithm}
\usepackage{algpseudocode}

\usepackage[colorlinks,breaklinks=true,bookmarks=false]{hyperref}

\iccvfinalcopy % *** Uncomment this line for the final submission

\def\iccvPaperID{9247} % *** Enter the ICCV Paper ID here

\ificcvfinal\fi

\begin{document}

%%%%%%%%% TITLE
\title{Towards Trustworthy Dataset Distillation}

\author{Shijie Ma\textsuperscript{1,2}, Fei Zhu\textsuperscript{1,2}, Zhen Cheng\textsuperscript{1,2}, Xu-Yao Zhang\textsuperscript{1,2}\thanks{Corresponding author.}
\and
\textsuperscript{1}MAIS, Institute of Automation, Chinese Academy of Sciences, Beijing 100190, China\\
\textsuperscript{2}School of Artificial Intelligence, University of Chinese Academy of Sciences, Beijing, 100049, China\\
{\tt\small \{mashijie2021, zhufei2018, chengzhen2019\}@ia.ac.cn}, {\tt\small xyz@nlpr.ia.ac.cn}
}

\maketitle
% Remove page # from the first page of camera-ready.
\ificcvfinal\thispagestyle{empty}\fi

%%%%%%%%% ABSTRACT
\begin{abstract}
   Efficiency and trustworthiness are two eternal pursuits when applying deep learning in real-world applications. With regard to efficiency, dataset distillation (DD) endeavors to reduce training costs by distilling the large dataset into a tiny synthetic dataset. However, existing methods merely concentrate on in-distribution (InD) classification in a closed-world setting, disregarding out-of-distribution (OOD) samples. On the other hand, OOD detection aims to enhance models' trustworthiness, which is always inefficiently achieved in full-data settings. For the first time, we simultaneously consider both issues and propose a novel paradigm called Trustworthy Dataset Distillation (TrustDD). By distilling both InD samples and outliers, the condensed datasets are capable of training models competent in both InD classification and OOD detection. To alleviate the requirement of real outlier data, we further propose to corrupt InD samples to generate pseudo-outliers, namely Pseudo-Outlier Exposure (POE). Comprehensive experiments on various settings demonstrate the effectiveness of TrustDD, and POE surpasses the state-of-the-art method Outlier Exposure (OE). Compared with the preceding DD, TrustDD is more trustworthy and applicable to open-world scenarios. Our code is available at \url{https://github.com/mashijie1028/TrustDD}
\end{abstract}

%%%%%%%%% BODY TEXT
\section{Introduction}

When applying algorithms and deploying models in practical scenarios, efficiency and trustworthiness are two crucial factors that require careful consideration. On the one hand, the large amounts of training data~\cite{5206848,radford2021learning} and computational resources are impractical and not affordable in downstream applications. On the other hand, trustworthiness and reliability also matter. One could not expect the test data to always be drawn from the same distribution as the training data, so deep networks are supposed to detect \emph{out-of-distribution} (OOD) samples~\cite{hendrycks2017a,hendrycks2018deep,liang2018enhancing} from unknown classes, rather than irresponsibly classify them into known categories, which may lead to catastrophic damage in safety-critical scenarios. For instance, in autonomous driving~\cite{bogdoll2022anomaly}, it is desirable to transfer control to the driver once the system detects anomalous situations rather than making arbitrary decisions. In medical imaging~\cite{10.1007/978-3-031-16452-1_4}, detecting abnormalities areas and unknown diseases are important for diagnosis.

\begin{figure}[!tb]
    \centering
    \includegraphics[width=.98\linewidth]{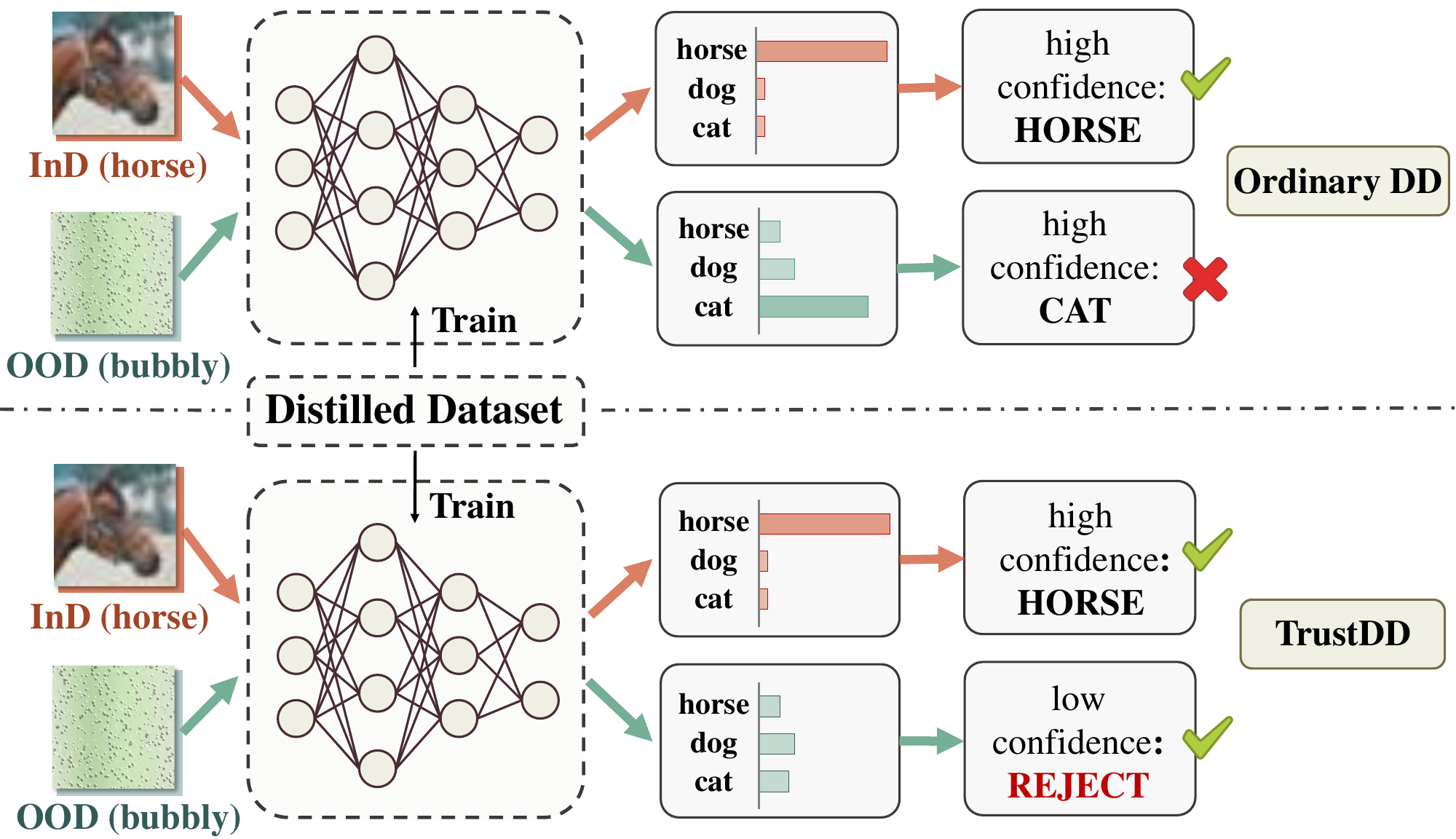}
    \vspace{-5pt}
    \caption{Advantages of the proposed TrustDD over preceding dataset distillation (Ordinary DD). For test OOD samples, models trained by Ordinary DD assign high confidence and misclassify bubbly samples in Texture~\cite{cimpoi2014describing} as cats, while TrustDD is capable to train reliable models to reject OOD samples with low confidence.}
    \label{fig:dd-trustdd-task}
    \vspace{-10pt}
\end{figure}

In terms of efficiency, efficient deep learning~\cite{menghani2021efficient} has emerged to reduce computational and training data requirements while minimizing performance degradation in various perspectives. Knowledge distillation~\cite{hinton2015distilling}, model quantization~\cite{jacob2018quantization}, and lightweight neural networks~\cite{howard2017mobilenets} offer efficient model options. From an orthogonal perspective, making data itself more efficient is also promising. \emph{Dataset Distillation} (DD)~\cite{wang2018dataset} aims to learn a small synthetic dataset, upon which model trained could gain similar performance on test dataset compared to the model trained on the original training dataset. Zhao \etal~\cite{zhao2021DSA,zhao2021DC} reformulated DD as a gradient matching problem to bypass the complex optimization procedure. Cazenavette \etal~\cite{cazenavette2022distillation} proposed to match training trajectories of the target network with a synthetically trained network to distill datasets, their method surpassed the former state-of-the-art (SOTA) by a large margin. Another line of research~\cite{loo2022efficient,nguyen2021dataset,NEURIPS2021_299a23a2,zhou2022dataset} resorts to kernel-based methods to obtain the closed-form solution in the inner loop. However, trustworthiness and reliability haven't been taken into account in dataset distillation studies, as a result, models tend to suffer from overconfidence problems.

As for the trustworthiness, OOD detection works~\cite{hendrycks2017a,lee2018simple,liang2018enhancing,liu2020energy} aim to improve models' ability to detect and reject OOD samples. Existing works are always conducted in full-data settings, which is inefficient and time-consuming.

\begin{figure}
    \centering
    \begin{subfigure}{.48\linewidth}
        \includegraphics[width=\linewidth]{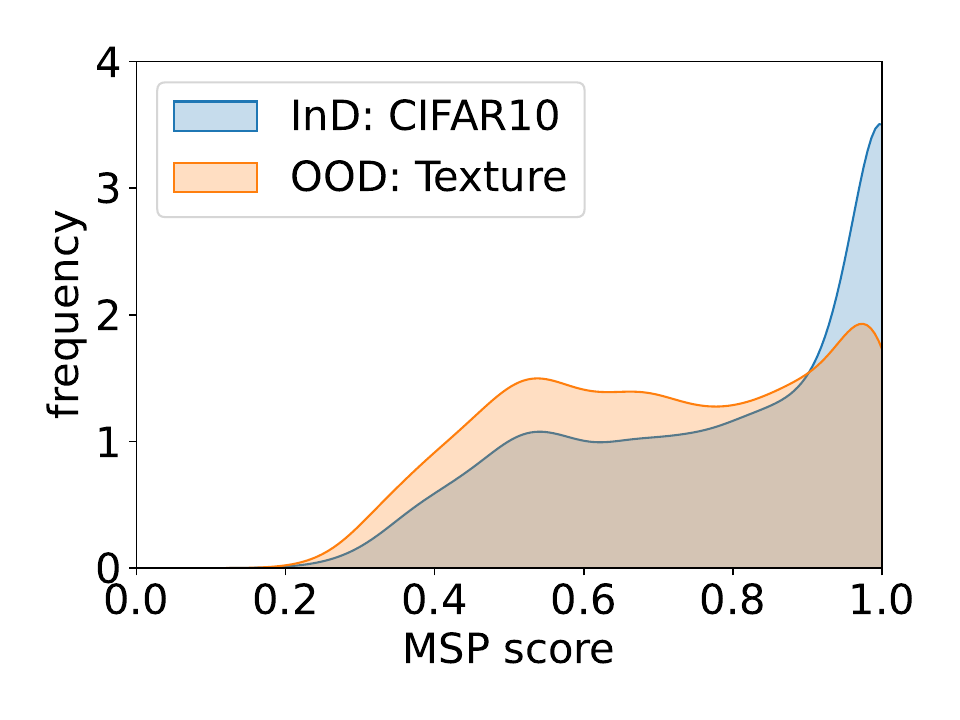}
        \vspace{-15pt}
        \caption{DD (FPR: 91.62)}
        \label{subfig:texture-baseline}
    \end{subfigure}
    \hfill
    \begin{subfigure}{.48\linewidth}
        \includegraphics[width=\linewidth]{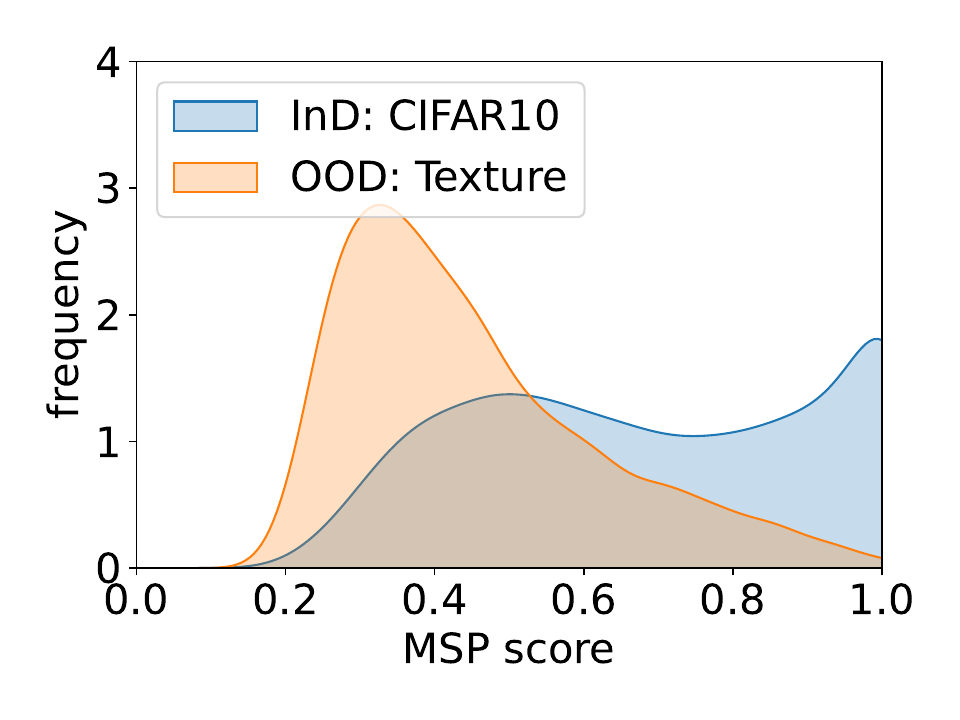}
        \vspace{-15pt}
        \caption{TrustDD (FPR: 75.49)}
        \label{subfig:texture-poe}
    \end{subfigure}
    \vspace{-10pt}
    \caption{Maximum Softmax Probability score distribution of InD and OOD samples. TrustDD could train better OOD detectors than ordinary DD (Figure~\ref{subfig:texture-baseline}$\to$~\ref{subfig:texture-poe}). Models are trained on CIFAR10 with 50 Image Per Class (IPC).}
    \label{fig:trustdd-imporve-ood}
    \vspace{-10pt}
\end{figure}

In short, although both efficiency and trustworthiness are important when deploying models in real-world applications, previous works solely focus on one aspect, which is less applicable. Motivated by this, in this paper, we consider both efficiency and trustworthiness simultaneously from the data perspective, and take the first step towards trustworthy dataset distillation. To achieve this goal, we take the spirit of Outlier Exposure (OE)~\cite{hendrycks2018deep} and establish a learning paradigm called \emph{Trustworthy Dataset Distillation} (TrustDD). To relax the requirement to auxiliary outlier dataset, we further propose to generate \emph{pseudo-outliers} from in-distribution samples and expose them to deep networks to boost OOD detection, which is named \emph{Pseudo-Outlier Exposure} (POE). We distill both InD and the generated outliers into a tiny dataset, such distilled dataset is capable of training not only competent classifiers but also reliable OOD detectors (See Figure~\ref{fig:trustdd-imporve-ood}.).

To the best of our knowledge, this is the first time that trustworthiness is taken into consideration in dataset distillation. Models trained on the distilled dataset perform well on in-distribution (InD) classification and out-of-distribution (OOD) detection simultaneously. Figure~\ref{fig:dd-trustdd-task} illustrates the superiority of TrustDD for handling both tasks.

Our main contributions are summarized as follows:
\vspace{-5pt}
\begin{itemize}
    \item We propose a novel paradigm, named \emph{Trustworthy Dataset Distillation} (TrustDD), considering both InD classification and OOD detection, which ensures both efficiency and trustworthiness and improves the OOD detection performance of DD.
    \vspace{-3pt}
    \item The proposed \emph{Pseudo-Outlier Exposure} (POE) in TrustDD performs comparably or surpasses the commonly-used SOTA OE even though POE does not need to collect real outlier data.
    \vspace{-3pt}
    \item Extensive experiments show TrustDD improves OOD detection performance by a large margin without the loss of InD classification accuracy, making DD trustworthy and applicable to the open-world.
\end{itemize}

\section{Related Work}

\subsection{Dataset Distillation}

Coreset selection~\cite{aljundi2019gradient,sener2018active} is an early-stage research in data-efficient learning. Most methods rely heavily on heuristics to select representatives, which are short-sighted and sub-optimal. \emph{Dataset Distillation} (DD)~\cite{sachdeva2023data,yu2023dataset}, also known as \emph{Dataset Condensation} (DC)~\cite{zhao2021DSA,zhao2021DC}, aims to synthesize a tiny dataset to train models which could perform comparably to the complete dataset setting. A pioneer work by Wang \etal~\cite{wang2018dataset} formulates DD in a bi-level meta-learning framework. One line of research focuses on simplifying the inner loop by solving the closed-form solution towards kernel ridge regression~\cite{loo2022efficient,nguyen2021dataset,NEURIPS2021_299a23a2,zhou2022dataset}. Alternatively, Zhao \etal~\cite{zhao2021DC} argued that making parameters trained on condensed data approximate the target parameters and derived gradient matching objective, which simplified DD from the parameter perspective. They further applied Differentiable Siamese Augmentation (DSA)~\cite{zhao2021DSA} to distill more informative images by enabling effective data augmentation on synthetic data. By combining meta-learning and parameter matching, Cazenavette \etal~\cite{cazenavette2022distillation} proposed directly Matching Training Trajectories (MTT) and achieved SOTA results. A recent work called TESLA~\cite{cui2022scaling} reduced GPU memory consumption and could be viewed as a memory-efficient version of MTT~\cite{cazenavette2022distillation}. Instead of matching gradients or parameters, recent works proposed to condense datasets by matching features~\cite{wang2022cafe,zhao2023DM}.

DD has been applied in many down-stream tasks, including continual learning~\cite{zhao2023DM}, privacy-preserving learning~\cite{pmlr-v162-dong22c} and neural architecture search~\cite{pmlr-v119-such20a,zhao2021DSA}. Generally speaking, existing DD algorithms focus solely on improving the accuracy of models trained on the distilled data, regardless of trustworthiness and reliability.

\subsection{Out-of-distribution Detection}

OOD detection~\cite{hendrycks2017a} seeks to detect samples from novel classes by assigning a confidence score, \ie, an indicator of \emph{normality}, for each sample. If the score is higher than a pre-defined threshold then such input is noted as in-distribution. A closely-related area to OOD detection is Open Set Recognition~\cite{9864101,vaze2022openset}, which focuses on known-class classification and unknown-class rejection. In this paper, we adopt the procedure of OOD detection for comprehensive evaluation.

Preceding works~\cite{hendrycks2017a,lee2018simple,liang2018enhancing,liu2020energy} of OOD detection concentrated on design suitable detection score function, \eg, Maximum Softmax Probability (MSP), Energy Score (Energy), Maximum Logit Score (MLS), to make InD and OOD samples more separable without interfering the training process. Following studies explored to learn better representations ~\cite{hendrycks2019using,tack2020csi}. Besides, Hendrycks \etal~\cite{hendrycks2018deep} introduced Outlier Exposure (OE) to explicitly utilize an auxiliary outlier dataset. OE is a simple yet effective method to achieve SOTA OOD detection performance.

However, OOD detection has largely overlooked scenarios with limited data, and DD is mainly conducted in closed-world. In this paper, we improve the OOD detection performance of DD to make DD both efficient and trustworthy, which is more applicable in real open-world.

\section{Background}

\subsection{Matching-based Dataset Distillation}
\vspace{-5pt}
\paragraph{Basic Notations.}
Let $\mathcal{T}=\{(\boldsymbol{x}^i,y^i)\}\big\vert_{i=1}^{|\mathcal{T}|}$ denote the whole large dataset, DD aims to distill the original dataset $\mathcal{T}$ into a tiny yet informative dataset $\mathcal{S}=\{(\boldsymbol{s}^i,y^i)\}\big\vert_{i=1}^{|\mathcal{S}|}$ \ie, $|\mathcal{S}|\ll |\mathcal{T}|$ so that model trained on $\mathcal{S}$ could exhibit comparable performance with the one trained on $\mathcal{T}$. $f$ denotes the neural network parameterized with $\theta$ and $f_\theta(\boldsymbol{x})$ is the output softmax probabilities on sample $\boldsymbol{x}$.

\vspace{-13pt}
\paragraph{Formulation.}
From the perspective of information theory, as in~\cite{pmlr-v162-lee22b}, the general objective of DD is as follows:
\vspace{-7pt}
\begin{equation}
    \mathcal{S}^\star=\arg\max_{\mathcal{S}}I(\mathcal{T};\mathcal{S}|\tau)
    \label{eq:mutual-information}
    \vspace{-5pt}
\end{equation}
where $I(\mathcal{T};\mathcal{S}|\tau)$ denotes the mutual information conditioned on the \emph{task indicator random variable} $\tau$. Maximizing the mutual information ensures the condensed dataset $\mathcal{S}^\star$ contains as much information in $\mathcal{T}$ as possible to train models to implement task $\tau$. For existing DD works~\cite{cazenavette2022distillation,cui2022scaling,pmlr-v162-lee22b,zhao2021DSA,zhao2023DM,zhao2021DC}, $\tau$ denotes InD classification task.

However, Eq.~\eqref{eq:mutual-information} only provides a general and implicit criterion. To be more task-specific and make DD computational tractable, prior works~\cite{wang2018dataset,zhao2021DC} derived Eq.~\eqref{eq:mutual-information} into the loss function form as follows:
\vspace{-5pt}
\begin{equation}
    \begin{split}
    &\mathcal{S}^\star=\arg\min_{\mathcal{S}} \mathcal{L}_{ce}^\mathcal{T}(\theta^\mathcal{S}(\mathcal{S}))\\
    \textrm{s.t.}\quad &\theta^\mathcal{S}(\mathcal{S})=\arg\min_{\theta}\mathcal{L}_{ce}^\mathcal{S}(\theta)
    \end{split}
    \label{eq:dd-loss}
    \vspace{-15pt}
\end{equation}
Eq.~\eqref{eq:dd-loss} is a meta-learning framework requiring bi-level optimization. In the inner loop, one could optimize the network parameters $\theta^\mathcal{S}$ on the distilled dataset $\mathcal{S}$, hence, $\theta^\mathcal{S}$ is a function of $\mathcal{S}$, while the outer loop learns to update $\mathcal{S}$ through minimizing the loss $\mathcal{L}_\textrm{ce}^\mathcal{T}$ on the real dataset. $\mathcal{L}_\textrm{ce}^\mathcal{T}$ and $\mathcal{L}_\textrm{ce}^\mathcal{S}$ are cross-entropy function on $\mathcal{T}$ and $\mathcal{S}$ respectively.

\vspace{-10pt}
\paragraph{Methods Review.}
We briefly review two parameters matching-based DD methods: DSA~\cite{zhao2021DSA} and MTT~\cite{cazenavette2022distillation}.
Zhao \etal~\cite{zhao2021DSA,zhao2021DC} proposed to circumvent the bi-level optimization problem in Eq.~\eqref{eq:dd-loss} by gradient matching. They contended that similar performance could be obtained by similar parameters, \ie, making $\theta^\mathcal{S}\approx\theta^\mathcal{T}$. With the same initialization $\theta_0^\mathcal{S}$ and $\theta_0^\mathcal{T}$, they further assumed that $\theta_t^\mathcal{S}=\theta_t^\mathcal{T}=\hat\theta_t$ could be achieved at each iteration $t$, and simplified Eq.~\eqref{eq:dd-loss} into a gradient matching problem:
\begin{equation}
    \mathcal{L}_\textrm{distill}(\mathcal{S})=D\big(\nabla_\theta\mathcal{L}(\mathcal{A}(\mathcal{S}),\hat\theta_t),\nabla_\theta\mathcal{L}(\mathcal{A}(\mathcal{T}),\hat\theta_t)\big)
    \label{eq:distill-loss-dsa}
\end{equation}
where $\mathcal{A}(\cdot)$ is the differentiable siamese augmentation.
Cazenavette \etal~\cite{cazenavette2022distillation} introduced MTT to directly align model parameters through trajectories matching:
\vspace{-5pt}
\begin{equation}
    \mathcal{L}_\textrm{distill}(\mathcal{S})=\frac{\Vert \hat\theta_{t+N}-\theta^\star_{t+M}\Vert_2^2}{\Vert\theta_t^\star-\theta_{t+M}^\star\Vert_2^2}
    \label{eq:distill-loss-mtt}
    \vspace{-5pt}
\end{equation}
where $\theta_t^\star$ and $\hat \theta_t$ denote parameters trained on $\mathcal{T}$ ($M$ updates in total) and $\mathcal{S}$ ($N$ updates in total) at time step $t$ respectively, and $N\ll M$.

Here, we put both DSA~\cite{zhao2021DSA} and MTT~\cite{cazenavette2022distillation} into a unified framework. The parameter update process is as follows:
\vspace{-5pt}
\begin{equation}
    \hat\theta_{i+1}=\hat\theta_{i}-\alpha \nabla_\theta \mathcal{L}(\mathcal{A}(\mathcal{S});\hat\theta_{i})
    \label{eq:iter-theta}
    \vspace{-5pt}
\end{equation}
In their cases, loss functions $\mathcal{L}$ in Eq.~\eqref{eq:distill-loss-dsa} and Eq.~\eqref{eq:iter-theta} are typically cross-entropy classification loss $\mathcal{L}_{ce}$. From Eq.~\eqref{eq:iter-theta} we can observe that both $\hat\theta_t$ and $\hat\theta_{t+N}$ in Eq.~\eqref{eq:distill-loss-dsa} and Eq.~\eqref{eq:distill-loss-mtt} are functions of $\mathcal{S}$, as a result, $\mathcal{L}_\textrm{distill}$ is also a function of $\mathcal{S}$ in both equations. One could optimize distillation loss $\mathcal{L}_\textrm{distill}$ by gradient descent to learn distilled dataset $\mathcal{S}$:
\vspace{-5pt}
\begin{equation}
    \mathcal{S}^\star=\arg\min_{\mathcal{S}}\mathcal{L}_\textrm{distill}(\mathcal{S})
    \label{eq:min-distill-loss}
\end{equation}

\subsection{Outlier Exposure} \label{subsec:oe}

Outlier Exposure (OE)~\cite{hendrycks2018deep} utilizes an auxiliary outlier dataset to improve OOD detection as follows:
\begin{equation}
    \mathcal{L}=\mathbb{E}_{(\boldsymbol{x},y)\sim \mathcal{D}_\textrm{in}}\mathcal{L}_\textrm{ce}(f_\theta(\boldsymbol{x}),y)+\lambda \mathbb{E}_{\boldsymbol{x}^\prime\sim \mathcal{D}_\textrm{out}}H(\mathcal{U};f_\theta(\boldsymbol{x}^\prime))
    \label{eq:loss-id-ood}
\end{equation}
where $\mathcal{L}_{ce}(f_\theta(x),y)$ is the cross-entropy loss for classification. For OOD detection, it is appropriate to minimize $H(\mathcal{U};f_\theta(\boldsymbol{x}^\prime))$, \ie, the cross-entropy between a uniform distribution and output probabilities, namely maximize the output uncertainty of outliers, which could in turn implicitly make the model generalize to detect unseen OOD samples. Note that the auxiliary dataset $\mathcal{D}_\textrm{out}$ should not overlap with test OOD datasets $\mathcal{D}_\textrm{out}^\textrm{test}$ for fair evaluation.

Similar to DD, OE~\cite{hendrycks2018deep} is also from the perspective of data independent of model structure and training tricks. Besides, OE serves as the SOTA method in OOD detection.

\section{Methodology}

\subsection{TrustDD Paradigm} \label{subsec:trustdd-paradigm}

To make DD trustworthy, we set \emph{task indicator variable} $\tau=(\tau_\textrm{InD};\tau_\textrm{OOD})$ in Eq.~\eqref{eq:mutual-information}, taking both InD classification and OOD detection into consideration. We expect such trustworthy DD to generalize well across various network architectures and OOD detection scores. Hence, similar to DD, we derive \emph{Trustworthy Dataset Distillation} (TrustDD) from the perspective of data and resort to OE~\cite{hendrycks2018deep}, \ie, utilizing outliers to boost OOD detection.

\vspace{-10pt}
\paragraph{TrustDD Notations.}
We additionally consider auxiliary outliers like OE~\cite{hendrycks2018deep} in TrustDD. Let $\mathcal{T}=\mathcal{T}_{\textrm{in}}\bigcup \mathcal{T}_\textrm{out}$ denote the full dataset, where $\mathcal{T}_\textrm{in}=\{(\boldsymbol{x}_\textrm{in}^i,y_\textrm{in}^i)\}\big\vert_{i=1}^{|\mathcal{T}_\textrm{in}|}$, $\mathcal{T}_\textrm{out}=\{\boldsymbol{x}_\textrm{out}^i\}\big\vert_{i=1}^{|\mathcal{T}_\textrm{out}|}$ are in-distribution (InD) and out-of-distribution (OOD) dataset (with no labels), respectively. Here, $\boldsymbol{x}_\textrm{in}^i\in\mathcal{X}_\textrm{in},y_\textrm{in}^i\in\{0,1,\cdots,C-1\},\boldsymbol{x}_\textrm{out}^i\in\mathcal{X}_\textrm{out}$. We propose to distill both InD and outliers into $\mathcal{S}=\mathcal{S}_\textrm{in}\bigcup \mathcal{S}_\textrm{out}$ where $\mathcal{S}_\textrm{in}=\{(\boldsymbol{s}_\textrm{in}^i,y_\textrm{in}^i)\}\big\vert_{i=1}^{|\mathcal{S}_\textrm{in}|}$, $\mathcal{S}_\textrm{out}=\{\boldsymbol{s}_\textrm{out}^i\}\big\vert_{i=1}^{|\mathcal{S}_\textrm{out}|}$ and $|\mathcal{S}|=|\mathcal{S}_\textrm{in}|+|\mathcal{S}_\textrm{out}| \ll |\mathcal{T}|=|\mathcal{T}_\textrm{in}|+|\mathcal{T}_\textrm{out}|$.

\vspace{-10pt}
\paragraph{TrustDD Formulation.}
Assume the true distribution of InD and OOD are $P_\textrm{in}$ and $P_\textrm{out}$ respectively. $\mathcal{R}=\mathcal{R}_\textrm{in}+\lambda\mathcal{R}_\textrm{out}$ denotes the integrated risk of InD classification risk and OOD detection risk, which could be rewritten as:
\begin{equation}
\vspace{-7pt}
    \begin{split}
        \mathcal{R}=&\mathbb{E}_{(\boldsymbol{x},y)\sim P_\textrm{in}}[\mathbbm{1}(\arg\max f_\theta(\boldsymbol{x})\neq y)]\\
        & +\lambda \mathbb{E}_{(\boldsymbol{x},y)\sim P_\textrm{in}}[\mathbbm{1}(S(\boldsymbol{x})<\delta)]\\
        & +\lambda \mathbb{E}_{\boldsymbol{x}^\prime\sim P_\textrm{out}}[\mathbbm{1}(S(\boldsymbol{x}^\prime)>\delta)]
    \end{split}
    \label{eq:trustdd-risk}
    \vspace{-5pt}
\end{equation}
 Here, $\mathbbm{1}(\cdot)$ is the indicator function and $S(\boldsymbol{x})$ denotes the detection score function, when $S(\boldsymbol{x})$ is larger than a pre-defined threshold $\delta$, $\boldsymbol{x}$ is referred to as positive (in-distribution). $\lambda$ is a trade-off between two tasks. 

In order to learn and update $\mathcal{S}$, similar to Eq.~\eqref{eq:dd-loss}, one can rewrite the risk in terms of the loss function. As a consequence, TrustDD paradigm could be written as follows:
\vspace{-5pt}
\begin{equation}
    \begin{split}
    &\mathcal{S}^\star=\arg\min_{\mathcal{S}} \mathcal{L}^\mathcal{T}(\theta^\mathcal{S}(\mathcal{S}))\\
    \textrm{s.t.}\quad &\theta^\mathcal{S}(\mathcal{S})=\arg\min_{\theta}\mathcal{L}^\mathcal{S}(\theta)
    \end{split}
    \label{eq:trustdd-formulation}
\end{equation}
\vspace{-5pt}
where:
\vspace{-2pt}
\begin{equation}
    \begin{split}
        \mathcal{L}^\mathcal{T}(\theta)=&\frac{1}{|\mathcal{T}_\textrm{in}|}\sum_{(\boldsymbol{x},y)\in \mathcal{T}_\textrm{in}}\mathcal{L}_{ce}(f_\theta(\boldsymbol{x}),y)\\
        &+\lambda \frac{1}{|\mathcal{T}_\textrm{out}|}\sum_{\boldsymbol{x}^\prime\in\mathcal{T}_\textrm{out}}H(\mathcal{U};f_\theta(\boldsymbol{x}^\prime))
    \end{split}
    \label{eq:loss-on-whole-set}
\end{equation}
\vspace{-8pt}
\begin{equation}
    \begin{split}
        \mathcal{L}^\mathcal{S}(\theta)=&\frac{1}{|\mathcal{S}_\textrm{in}|}\sum_{(\boldsymbol{s},y)\in \mathcal{S}_\textrm{in}}\mathcal{L}_{ce}(f_\theta(\boldsymbol{s}),y)\\
        &+\lambda \frac{1}{|\mathcal{S}_\textrm{out}|}\sum_{\boldsymbol{s}^\prime\in\mathcal{S}_\textrm{out}}H(\mathcal{U};f_\theta(\boldsymbol{s}^\prime))
    \end{split}
    \label{eq:loss-on-distilled-set}
\end{equation}
$\mathcal{L}^\mathcal{T}$ and $\mathcal{L}^\mathcal{S}$ are loss functions on $\mathcal{T}$ and $\mathcal{S}$ respectively. Intuitively, Eq.~\eqref{eq:trustdd-formulation} ensures model $\theta^\mathcal{S}$ trained on $\mathcal{S}$ could perform well on $\mathcal{T}$ considering both InD and OOD tasks, as a result, we distill $\mathcal{T}$ into a tiny yet informative dataset $\mathcal{S}^\star$.

In fact, the formulation of TrustDD in Eq.~\eqref{eq:trustdd-formulation} is the same as Eq.~\eqref{eq:dd-loss} when $\lambda=0$. In the case of TrustDD, $\lambda>0$, which means that we consider both InD classification and OOD detection, while prior DD works~\cite{wang2018dataset,zhao2021DSA,zhao2021DC} only consider the first term in Eq.~\eqref{eq:loss-on-whole-set} and Eq.~\eqref{eq:loss-on-distilled-set}.

\subsection{Pseudo-Outlier Exposure} \label{subsec:poe}
\paragraph{Motivation.}
OE~\cite{hendrycks2018deep} has a clear limitation as it relies on curated auxiliary data. Nevertheless, it is not always practical to collect such outliers. Besides, OE~\cite{hendrycks2018deep} is highly sensitive to the outlier data, as seen in Table~\ref{tab:oe-outlier-type}. When choosing random noises or a specific OOD dataset as auxiliary outliers, the performance gain brought by OE is very subtle. As a result, employing a curated outlier dataset (\eg, 300K Random Images~\cite{hendrycks2018deep}) is essential for OE. To avoid the dependence on real outlier data and make OOD detection more applicable, we propose to synthesize \emph{pseudo-outliers} from in-distribution data, \ie, the original dataset. We argue that in-distribution corruption is effective to generate a large number of \emph{pseudo-outliers}, which can be used to maximize the output uncertainty in Eq.~\eqref{eq:loss-id-ood}. We refer to this method as \emph{Pseudo-Outlier Exposure} (POE). POE is more practical and applicable than OE without the requirement of real outliers.

\begin{table}[!h]
\vspace{-5pt}
\centering
\scriptsize
\caption{OOD performance of models trained with IPC=10 when applying different auxiliary outliers in OE~\cite{hendrycks2018deep}. By default, OE~\cite{hendrycks2018deep} adopts 300K Random Images (300K).}
\label{tab:oe-outlier-type}
\vspace{-7pt}
\resizebox{.85\linewidth}{!}{
\begin{tabular}{@{}cccccc@{}}
\toprule
Outlier Data & None & Gauss & Uniform & SVHN & 300K \\ \midrule
AUROC & 63.36 & 66.81 & 66.71 & 66.33 & \textbf{70.12} \\
AUPR-IN & 67.45 & 70.59 & 70.13 & 69.90 & \textbf{73.51} \\ \bottomrule
\end{tabular}
}
\vspace{-10pt}
\end{table}

\begin{figure}[t]
     \centering
     \begin{subfigure}{0.18\linewidth}
         \centering
         \includegraphics[width=\linewidth]{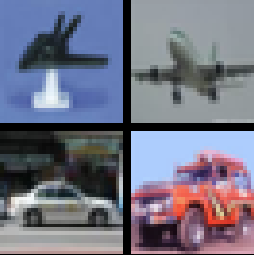}
         \subcaption*{\texttt{original}}
     \end{subfigure}
     \hfill
     \begin{subfigure}{0.18\linewidth}
         \centering
         \includegraphics[width=\linewidth]{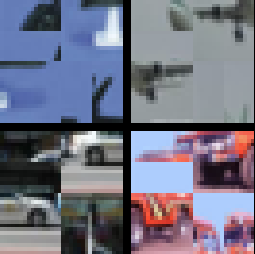}
         \subcaption*{\texttt{jigsaw}}
     \end{subfigure}
     \hfill
     \begin{subfigure}{0.18\linewidth}
         \centering
         \includegraphics[width=\linewidth]{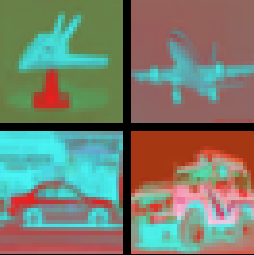}
         \subcaption*{\texttt{invert}}
     \end{subfigure}
     \hfill
     \begin{subfigure}{0.18\linewidth}
         \centering
         \includegraphics[width=\linewidth]{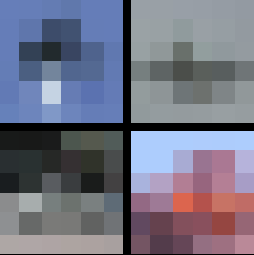}
         \subcaption*{\texttt{mosaic}}
     \end{subfigure}
     \hfill
     \begin{subfigure}{0.18\linewidth}
         \centering
         \includegraphics[width=\linewidth]{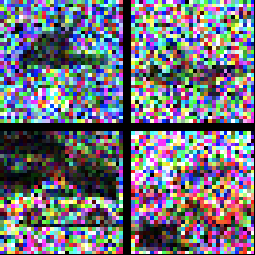}
         \subcaption*{\texttt{speckle}}
     \end{subfigure}
     \vspace{-8pt}
        \caption{Visualization of InD corruption to synthesize \emph{pseudo-outliers} on CIFAR10~\cite{krizhevsky2009learning}. The corruption transformations are: \texttt{jigsaw}, \texttt{invert}, \texttt{mosaic} and \texttt{speckle}.}
        \label{fig:cifar-corrupt}
    \vspace{-10pt}
\end{figure}

\vspace{-10pt}
\paragraph{Corruption Transformations.}
OOD samples refer to the ones exhibiting semantic shifts, namely belonging to categories outside of the training data. To synthesize such outliers, one should ensure that corruption leads to a noticeable semantic shift. If the corruption is semantic-preserving, then \emph{pseudo-outliers} are also in-distribution samples, minimizing $H(\mathcal{U};f_\theta(\boldsymbol{x}))$ in Eq.~\eqref{eq:loss-id-ood} could degrade the InD classification performance, which leads to bad OOD detection~\cite{vaze2022openset}. For natural scene images like CIFAR~\cite{krizhevsky2009learning} and ImageNet~\cite{5206848}, we mainly perform four corruption transformations: \texttt{jigsaw}, \texttt{invert}, \texttt{mosaic} and \texttt{speckle}. An example of \emph{pseudo-outliers} is shown in Figure~\ref{fig:cifar-corrupt}.

Regarding the specific operations, $\texttt{jigsaw}$ divides InD images $\boldsymbol{x}_\textrm{in}\in\mathbb{R}^{C\times H\times W}$ into $6\sim8$ patches and shuffle them to generate $\boldsymbol{x}_\textrm{out}$. $\texttt{invert}$ means channel-wise inversion on certain channels, \ie, $\boldsymbol{x}_\textrm{out}[C,:,:]=1-\boldsymbol{x}_\textrm{in}[C,:,:]$. \texttt{mosaic} blurs $\boldsymbol{x}_\textrm{in}$ to make it unrecognizable. $\texttt{speckle}$ corrupts the input in a pixel-wise manner, \ie, $\boldsymbol{x}_\textrm{out}=\boldsymbol{x}_\textrm{in}+\boldsymbol{x}_\textrm{in}\times\texttt{rand\_like}(\boldsymbol{x}_\textrm{in})$ and clip it to range $[0,1]$.

\subsection{Overall Learning Framework} \label{subsec:overall-framework}
\vspace{-5pt}
Based on the TrustDD paradigm and POE, we can derive the overall learning framework of TrustDD in Algorithm~\ref{alg:trustdd-paradigm}.

\begin{algorithm}[t]
    \small
    \caption{Trustworthy Dataset Distillation (TrustDD)}
    \label{alg:trustdd-paradigm}
    \begin{algorithmic}[1]
        \Require Original dataset $\mathcal{T}_\textrm{in}$.
        \Require Corruption function $\mathcal{C}(\cdot,\cdot)$ with transformation set $\Omega=\{\texttt{jigsaw},\texttt{invert},\texttt{mosaic},\texttt{speckle},\cdots\}$, differentiable augmentation function $\mathcal{A}(\cdot)$.
        \Require  Network learning rate $\alpha_1$, distilled image learning rate $\alpha_2$.
        \Require Number of network updates $N$ and image updates $N_\mathcal{S}$ in each iteration.
        \Require Integrated loss $\mathcal{L}^\mathcal{T}(\theta)$ in Eq.~\eqref{eq:loss-on-whole-set} and $\mathcal{L}^\mathcal{S}(\theta)$ in Eq.~\eqref{eq:loss-on-distilled-set}, distillation loss $\mathcal{L}_\textrm{distill}(\mathcal{S})$. (like Eq.~\eqref{eq:distill-loss-dsa} and Eq.~\eqref{eq:distill-loss-mtt})
        \State $\triangleright$ Corrupt $\mathcal{T}_\textrm{in}$ to generate pseudo-outlier dataset:
        \State \quad\quad $\mathcal{T}_\textrm{out}=\mathcal{C}(\mathcal{T}_\textrm{in},\Omega)$ \label{step:corrupt}
        \State $\triangleright$ Initialize distilled dataset:
        \State \quad \quad $\mathcal{S}_\textrm{in}\sim \mathcal{T}_\textrm{in}$, $\mathcal{S}_\textrm{out}\sim \mathcal{T}_\textrm{out}$ and $\mathcal{S}=\mathcal{S}_\textrm{in}\bigcup \mathcal{S}_\textrm{out}$
        \While{not converged}
            \State $\triangleright$ Initialize network parameters: $\hat\theta_t$ (obtained from the previous iteration or randomly sampled) \label{step:initialize}
            \State $\triangleright$ Update network parameters $N$ times:
            \State \quad \quad $\hat\theta_{i+1}=\hat\theta_{i}-\alpha_1\nabla_\theta \mathcal{L}^\mathcal{S}(\hat\theta_i)$ \label{step:update-network}
            \State $\triangleright$ Compute distillation loss $\mathcal{L}_\textrm{distill}(\mathcal{S})$ on $\hat\theta_{t+N}$ \label{step:compute-distill-loss}
            \State $\triangleright$ Update distilled images $N_\mathcal{S}$ times:
            \State \quad \quad $\mathcal{S}_\textrm{in} \leftarrow \mathcal{S}_\textrm{in}-\alpha_2\nabla_{\mathcal{S}_\textrm{in}}\mathcal{L}_\textrm{distill}(\mathcal{S})$ \label{step:update-distill-in} 
            \State \quad \quad $\mathcal{S}_\textrm{out} \leftarrow \mathcal{S}_\textrm{out}-\alpha_2\nabla_{\mathcal{S}_\textrm{out}}\mathcal{L}_\textrm{distill}(\mathcal{S})$
            \label{step:update-distill-out}
        \EndWhile
        \Ensure {Distilled dataset $\mathcal{S}^\star=\mathcal{S}^\star_\textrm{in}\bigcup \mathcal{S}^\star_\textrm{out}$}
    \end{algorithmic}
\end{algorithm}

Firstly, we corrupt InD data to generate pseudo outliers $\mathcal{T}_\textrm{out}=\mathcal{C}(\mathcal{T}_\textrm{in},\Omega)$. By simply substituting the single cross-entropy loss $\mathcal{L}_\textrm{ce}$ with \emph{integrated loss} and distilling both InD and generated outliers, we could upgrade ordinary DD to TrustDD. For instance, one can simply add a loss term $H(\mathcal{U};f_\theta(\mathcal{S}_\textrm{out}))$ to $\mathcal{L}_\textrm{ce}$ in Eq.~\eqref{eq:distill-loss-dsa} and Eq.~\eqref{eq:iter-theta} so as to make DSA~\cite{zhao2021DSA} and MTT~\cite{cazenavette2022distillation} trustworthy and reliable. For DSA~\cite{zhao2021DSA}, Line~\ref{step:initialize} initializes $\hat\theta_t$ from the previous iteration while for MTT~\cite{cazenavette2022distillation} sampled from expert parameters $\theta^\star_t$. Note that we have implemented \emph{integrated loss} twice, $\mathcal{L}^\mathcal{T}$ (Eq.~\eqref{eq:loss-on-whole-set}) is to compute $\mathcal{L}_\textrm{distill}$ in Line~\ref{step:compute-distill-loss} and further update distilled images in Line~\ref{step:update-distill-in} and Line~\ref{step:update-distill-out}, $\mathcal{L}^\mathcal{S}$ (Eq.~\eqref{eq:loss-on-distilled-set}) is to update the network in Line~\ref{step:update-network}.

In general, TrustDD could be seamlessly built upon most matching-based DD methods~\cite{cazenavette2022distillation,cui2022scaling,pmlr-v162-lee22b,zhao2021DSA,zhao2021DC} by applying corresponding $\mathcal{L}_\textrm{distill}(\mathcal{S})$ in Algorithm~\ref{alg:trustdd-paradigm}. Besides, TrustDD could be equipped with OE or POE, by choosing real outliers or \emph{pseudo-outliers} as $\mathcal{T}_\textrm{out}$.

\section{Experiments} \label{sec:exp}

\subsection{Experimental Setup} \label{subsec:exp-setup}
\paragraph{Datasets.}
Following~\cite{cazenavette2022distillation,zhao2021DSA}, we evaluate TrustDD on various datasets, including natural scene datasets (CIFAR~\cite{krizhevsky2009learning} and ImageNet Subsets~\cite{cazenavette2022distillation,imagenette}) and digit datasets (MNIST~\cite{lecun1998gradient} and SVHN~\cite{37648}). The corresponding test datasets for OOD detection~\cite{yang2022openood} are shown in Table~\ref{tab:id-ood-list}. $\mathcal{D}_\textrm{in}$ and $\mathcal{D}_\textrm{out}^\textrm{test}$ do not have any category intersections. By default, the size ratio of test InD and OOD dataset is kept as $1:1$.

\vspace{-13pt}
\paragraph{Metrics.}
For OOD detection~\cite{hendrycks2017a}, we use the common metrics: (1) \emph{\textbf{FPR95}}: False Positive Rate when True Positive Rate is at $95\%$. (2) \emph{\textbf{AUROC}}: Area Under the Receiver Operating Characteristic curve. (3) \emph{\textbf{AUPR-IN}} and \emph{\textbf{AUPR-OUT}}: Area under the Precision-Recall curve where InD and OOD samples are regarded as positives respectively. 

By default, we adopt Maximum Softmax Probabilities (MSP)~\cite{hendrycks2017a} as the OOD detection score in our experiments. Other confidence scores can be easily applied in our method which will be shown in Section~\ref{subsec:main-results}.

\begin{table}[!t]
% \vspace{-5pt}
\setlength\tabcolsep{6pt}
\centering
\renewcommand{\arraystretch}{1}
\caption{In-distribution datasets $\mathcal{D}_\textrm{in}$ for InD classification and corresponding test out-of-distribution datasets $\mathcal{D}_\textrm{out}^\textrm{test}$.}
\label{tab:id-ood-list}
\vspace{-7pt}
\resizebox{.98\linewidth}{!}{
\begin{tabular}{cc}
\hline
InD datasets $\mathcal{D}_\textrm{in}$ & OOD datasets $\mathcal{D}_\textrm{out}^\textrm{test}$ \\ \hline
CIFAR~\cite{krizhevsky2009learning} & \begin{tabular}[c]{@{}c@{}}Texture~\cite{cimpoi2014describing}, Places365~\cite{zhou2017places},\\ Tiny ImageNet~\cite{tiny}, LSUN~\cite{yu2015lsun}, iSUN~\cite{xu2015turkergaze}\end{tabular} \\ \hline
ImageNet Subsets~\cite{cazenavette2022distillation,imagenette} & \begin{tabular}[c]{@{}c@{}}Texture~\cite{cimpoi2014describing}, Species~\cite{pmlr-v162-hendrycks22a}, iNaturalist~\cite{van2018inaturalist},\\ ImageNet-O~\cite{hendrycks2021natural}, OpenImage-O~\cite{wang2022vim}\end{tabular} \\ \hline
MNIST~\cite{lecun1998gradient} & \begin{tabular}[c]{@{}c@{}}Texture~\cite{cimpoi2014describing}, Places365~\cite{zhou2017places}, Tiny ImageNet~\cite{tiny},\\ notMNIST~\cite{notmnist}, FashionMNIST~\cite{xiao2017/online}, CIFAR10~\cite{krizhevsky2009learning}\end{tabular} \\ \hline
SVHN~\cite{37648} & \begin{tabular}[c]{@{}c@{}}Texture~\cite{cimpoi2014describing}, Places365~\cite{zhou2017places}, Tiny ImageNet~\cite{tiny},\\ LSUN~\cite{yu2015lsun}, iSUN~\cite{xu2015turkergaze}, CIFAR10~\cite{krizhevsky2009learning}\end{tabular} \\ \hline
\end{tabular}
}
\vspace{-13pt}
\end{table}

\vspace{-13pt}
\paragraph{Training and Evaluation Procedure.} The training and evaluation include three steps. (1) Distill the whole dataset into a small informative dataset, \ie, dataset distillation. By default, we fix the distilled InD size $|\mathcal{S}_\textrm{in}|=\#\texttt{class}\times \texttt{IPC}$, for baseline, we set $|\mathcal{S}_\textrm{out}|=0$, \ie without outlier distillation, while for TrustDD on OE and POE, we keep $|\mathcal{S}_\textrm{out}|=|\mathcal{S}_\textrm{in}|$. We also compare the baseline with OE and POE with the same total distilled size $|\mathcal{S}|$ in Section~\ref{subsec:further-analysis}. (2) Train models on the distilled dataset. (3) Evaluate the trained models on real InD datasets and real test OOD datasets. OOD detection results on gaussian and uniform noises are only reported in Table~\ref{tab:dsa-ood-cifar} and Table~\ref{tab:tesla-ood-cifar}, for other OOD-related results, we report the average over $\mathcal{D}_\textrm{out}^\textrm{test}$ datasets in Table~\ref{tab:id-ood-list}. In all experiments, we distill the dataset 3 times, 5 models are trained on each distilled dataset, and the average of all 15 evaluations is reported.

\vspace{-13pt}
\paragraph{DD Frameworks.} In this paper, we mainly implement TrustDD on two well-known algorithms: DSA~\cite{zhao2021DSA} and MTT~\cite{cazenavette2022distillation}. When the number of categories or Image Per Class (IPC) is high, we resort to TESLA~\cite{cui2022scaling}, a memory-efficient version of MTT. To be more specific, for CIFAR, we build TrustDD upon DSA and TESLA with IPC=10, we also explore IPC=1 and IPC=50 in Section~\ref{subsec:ablation}. For digit datasets (MNIST and SVHN), we implement DSA with IPC=10. For ImageNet subsets, we employ MTT with IPC=1. Besides, TrustDD could be implemented with both OE and POE. For OE, we instantiate $\mathcal{T}_\textrm{out}$ with 300K Random Images following~\cite{hendrycks2018deep}, for POE, we generate $\mathcal{T}_\textrm{out}$ via ensemble of InD corruption as Line~\ref{step:corrupt} in Algorithm~\ref{alg:trustdd-paradigm}.

\begin{table*}[!h]
\setlength\tabcolsep{4pt}
\centering
\renewcommand{\arraystretch}{0.9}
%\scriptsize
\caption{OOD detection performance of models trained on DSA distilled dataset on CIFAR, with IPC=10.}
\label{tab:dsa-ood-cifar}
\vspace{-10pt}
\resizebox{.9\textwidth}{!}{
\begin{tabular}{@{}cccccccccccccc@{}}
\toprule
\multirow{2}{*}{$\mathcal{D}_{\textrm{in}}$} & \multirow{2}{*}{$\mathcal{D}_\textrm{out}^\textrm{test}$} & \multicolumn{3}{c}{FPR95 $\downarrow$} & \multicolumn{3}{c}{AUROC $\uparrow$} & \multicolumn{3}{c}{AUPR-IN $\uparrow$} & \multicolumn{3}{c}{AUPR-OUT $\uparrow$} \\ \cmidrule(l){3-5} \cmidrule(l){6-8} \cmidrule(l){9-11} \cmidrule(l){12-14}
 &  & baseline & OE & POE (ours) & baseline & OE & POE (ours) & baseline & OE & POE (ours) & baseline & OE & POE (ours) \\ \midrule
\multirow{8}{*}{CIFAR10} & Gaussian & 68.62 & 69.59 & \textbf{28.70} & 83.82 & 85.37 & \textbf{95.30} & 86.40 & 88.07 & \textbf{96.03} & 80.75 & 81.45 & \textbf{94.52} \\
 & Uniform & 68.70 & 70.24 & \textbf{29.53} & 83.66 & 85.33 & \textbf{95.17} & 86.23 & 88.05 & \textbf{95.93} & 86.60 & 81.28 & \textbf{94.38} \\
 & Texture & 91.74 & 81.91 & \textbf{81.10} & 61.73 & \textbf{73.73} & 72.85 & 75.39 & \textbf{83.92} & 83.00 & 44.08 & \textbf{58.51} & 58.13 \\
 & Places365 & 90.44 & 86.93 & \textbf{86.79} & 62.98 & \textbf{68.82} & 68.62 & 64.58 & 69.86 & \textbf{70.44} & 59.88 & \textbf{65.60} & 65.46 \\
 & Tiny ImageNet & 90.17 & 87.44 & \textbf{82.46} & 63.44 & 68.98 & \textbf{71.82} & 64.83 & 70.79 & \textbf{72.74} & 60.35 & 65.36 & \textbf{69.49} \\
 & LSUN & 87.65 & 84.13 & \textbf{80.21} & 66.85 & 73.56 & \textbf{75.12} & 69.17 & 75.61 & \textbf{76.73} & 63.65 & 69.74 & \textbf{72.24} \\
 & iSUN & 88.61 & 85.66 & \textbf{80.67} & 65.28 & 71.42 & \textbf{74.10} & 69.62 & 75.05 & \textbf{77.23} & 59.51 & 65.28 & \textbf{69.20} \\ \cmidrule(l){2-14} 
 & \textbf{mean} & 83.70 & 80.84 & \textbf{67.07} & 69.68 & 75.31 & \textbf{79.00} & 73.75 & 78.76 & \textbf{81.73} & 64.97 & 69.60 & \textbf{74.77} \\ \midrule
\multirow{8}{*}{CIFAR100} & Gaussian & 98.84 & 88.90 & \textbf{1.82} & 36.77 & 72.57 & \textbf{99.56} & 49.41 & 77.81 & \textbf{99.60} & 41.36 & 65.64 & \textbf{99.53} \\
 & Uniform & 99.01 & 89.66 & \textbf{1.99} & 36.22 & 71.51 & \textbf{99.52} & 49.11 & 77.04 & \textbf{99.56} & 41.07 & 64.42 & \textbf{99.49} \\
 & Texture & 96.55 & 83.05 & \textbf{78.63} & 47.30 & 63.89 & \textbf{70.34} & 63.68 & 75.41 & \textbf{80.42} & 33.64 & 52.53 & \textbf{58.77} \\
 & Places365 & 91.91 & \textbf{87.28} & 89.02 & 59.33 & \textbf{63.94} & 63.18 & 60.08 & \textbf{63.72} & 63.62 & 57.19 & \textbf{62.31} & 61.05 \\
 & Tiny ImageNet & 91.79 & 87.87 & \textbf{71.45} & 61.67 & 66.95 & \textbf{75.93} & 64.23 & 68.93 & \textbf{75.95} & 58.51 & 63.42 & \textbf{76.18} \\
 & LSUN & 91.16 & 83.19 & \textbf{69.46} & 62.87 & 71.47 & \textbf{77.83} & 65.71 & 72.66 & \textbf{77.66} & 59.61 & 68.49 & \textbf{77.60} \\
 & iSUN & 91.66 & 85.70 & \textbf{68.93} & 61.50 & 68.01 & \textbf{76.01} & 66.02 & 71.13 & \textbf{76.77} & 55.85 & 62.77 & \textbf{75.09} \\ \cmidrule(l){2-14} 
 & \textbf{mean} & 94.42 & 86.52 & \textbf{54.47} & 52.24 & 68.33 & \textbf{80.34} & 59.75 & 72.39 & \textbf{81.94} & 49.60 & 62.80 & \textbf{78.24} \\ \bottomrule
\end{tabular}
}
\vspace{-5pt}
\end{table*}

\begin{table*}[!h]
\setlength\tabcolsep{4pt}
\centering
\renewcommand{\arraystretch}{0.9}
%\scriptsize
\caption{OOD detection performance of models trained on TESLA distilled dataset on CIFAR, with IPC=10.}
\label{tab:tesla-ood-cifar}
\vspace{-10pt}
\resizebox{.9\textwidth}{!}{
\begin{tabular}{@{}cccccccccccccc@{}}
\toprule
\multirow{2}{*}{$\mathcal{D}_{\textrm{in}}$} & \multirow{2}{*}{$\mathcal{D}_\textrm{out}^\textrm{test}$} & \multicolumn{3}{c}{FPR95 $\downarrow$} & \multicolumn{3}{c}{AUROC $\uparrow$} & \multicolumn{3}{c}{AUPR-IN $\uparrow$} & \multicolumn{3}{c}{AUPR-OUT $\uparrow$} \\ \cmidrule(l){3-5} \cmidrule(l){6-8} \cmidrule(l){9-11} \cmidrule(l){12-14}
 &  & baseline & OE & POE (ours) & baseline & OE & POE (ours) & baseline & OE & POE (ours) & baseline & OE & POE (ours) \\ \midrule
\multirow{8}{*}{CIFAR10} & Gaussian & 99.58 & 97.79 & \textbf{22.33} & 38.40 & 54.10 & \textbf{96.20} & 51.63 & 64.23 & \textbf{96.79} & 40.72 & 49.39 & \textbf{95.52} \\
 & Uniform & 99.61 & 98.08 & \textbf{22.93} & 37.48 & 53.03 & \textbf{96.09} & 51.03 & 63.48 & \textbf{96.68} & 40.33 & 48.61 & \textbf{95.43} \\
 & Texture & 91.23 & 84.64 & \textbf{83.99} & 61.97 & 67.42 & \textbf{70.10} & 75.14 & 78.12 & \textbf{80.99} & 44.96 & 52.74 & \textbf{54.45} \\
 & Places365 & 89.32 & \textbf{86.32} & 87.04 & 63.66 & \textbf{69.68} & 67.51 & 64.62 & \textbf{71.28} & 68.75 & 61.13 & \textbf{66.12} & 64.41 \\
 & Tiny ImageNet & 89.04 & 87.84 & \textbf{82.52} & 64.88 & 68.35 & \textbf{73.01} & 66.81 & 70.69 & \textbf{74.56} & 61.85 & 64.49 & \textbf{69.76} \\
 & LSUN & 87.18 & 85.28 & \textbf{81.95} & 67.17 & 71.55 & \textbf{74.44} & 69.01 & 74.01 & \textbf{76.47} & 64.38 & 67.57 & \textbf{70.72} \\
 & iSUN & 88.12 & 87.01 & \textbf{82.19} & 65.79 & 69.50 & \textbf{73.10} & 69.71 & 73.92 & \textbf{76.54} & 60.35 & 62.97 & \textbf{67.52} \\ \cmidrule(l){2-14} 
 & \textbf{mean} & 92.01 & 89.57 & \textbf{66.14} & 57.05 & 64.80 & \textbf{78.64} & 63.99 & 70.82 & \textbf{81.54} & 53.39 & 58.84 & \textbf{73.97} \\ \midrule
\multirow{8}{*}{CIFAR100} & Gaussian & 99.93 & 99.59 & \textbf{0.00} & 35.87 & 56.92 & \textbf{100.0} & 48.06 & 67.29 & \textbf{100.0} & 39.28 & 49.30 & \textbf{100.0} \\
 & Uniform & 99.93 & 99.74 & \textbf{0.00} & 35.55 & 56.00 & \textbf{100.0} & 48.04 & 66.67 & \textbf{100.0} & 39.14 & 48.68 & \textbf{100.0} \\
 & Texture & 96.10 & 88.98 & \textbf{74.67} & 46.42 & 61.59 & \textbf{74.00} & 61.42 & 75.04 & \textbf{82.90} & 33.64 & 46.68 & \textbf{63.21} \\
 & Places365 & 92.67 & 88.82 & \textbf{87.78} & 58.92 & 64.76 & \textbf{66.30} & 60.39 & 66.22 & \textbf{67.25} & 56.26 & 62.03 & \textbf{63.54} \\
 & Tiny ImageNet & 93.70 & 93.10 & \textbf{68.71} & 57.88 & 62.22 & \textbf{78.43} & 60.53 & 66.09 & \textbf{78.21} & 54.86 & 57.66 & \textbf{78.34} \\
 & LSUN & 94.17 & 93.79 & \textbf{71.29} & 57.91 & 62.59 & \textbf{78.64} & 60.96 & 67.28 & \textbf{79.44} & 54.47 & 57.31 & \textbf{77.40} \\
 & iSUN & 94.32 & 94.30 & \textbf{70.00} & 57.02 & 60.90 & \textbf{77.12} & 62.43 & 67.73 & \textbf{78.27} & 51.11 & 53.21 & \textbf{75.38} \\ \cmidrule(l){2-14} 
 & \textbf{mean} & 95.83 & 94.05 & \textbf{53.21} & 49.94 & 60.71 & \textbf{82.07} & 57.40 & 68.05 & \textbf{83.72} & 46.97 & 53.55 & \textbf{79.70} \\ \bottomrule
\end{tabular}
}
\vspace{-10pt}
\end{table*}

\vspace{-13pt}
\paragraph{Network Architectures and Hyper-parameters.}
We mainly implement TrustDD on ConvNet, which is default and canonical in the literature of DD~\cite{cazenavette2022distillation,cui2022scaling,zhao2021DSA}. For ImageNet subsets, we employ ConvNets with 5 convolutional layers, while 3 for the remaining datasets. For the trade-off weight, we simply set $\lambda=0.5$ in all experiments as in~\cite{hendrycks2018deep}. Other DD algorithm-related hyper-parameters are directly adopted from the corresponding papers~\cite{cazenavette2022distillation,cui2022scaling,zhao2021DSA}. For fair comparison among DD methods, we do not employ ZCA whitening for pre-processing in MTT~\cite{cazenavette2022distillation} and TESLA~\cite{cui2022scaling}, so the InD classification might degrade slightly compared with their official results, however, this does not hinder demonstrating the effectiveness of TrustDD through experiments.

\subsection{Main Results} \label{subsec:main-results}

% \paragraph{CIFAR~\cite{krizhevsky2009learning}.} For CIFAR10 and CIFAR100 dataset, we implement TrustDD upon DSA~\cite{zhao2021DSA} and TESLA~\cite{cui2022scaling} DD framework with IPC=10, the OOD detection results are shown in Table~\ref{tab:dsa-ood-cifar} and Table~\ref{tab:tesla-ood-cifar} respectively. Besides, we also conduct in-distribution classification experiments on CIFAR, results are shown in Figure~\ref{fig:id-acc}.

\paragraph{Ordinary DD performs poorly on OOD detection.}
Previous DD works~\cite{cazenavette2022distillation,cui2022scaling,zhao2021DSA} are confined to a closed world setting, serving as the baseline with very weak OOD detection performance. For instance, as shown in Figure~\ref{fig:trustdd-imporve-ood}, the model trained on dataset condensed from ordinary DD obtains $91.62\%$ FPR95, \ie, misclassifies $91.62\%$ negative (OOD) samples as InD when the true positive rate is $95\%$, while TrustDD reduces such error by $\sim16.13\%$.

\vspace{-13pt}
\paragraph{Is OOD knowledge distillable?}
Results in Table~\ref{tab:dsa-ood-cifar} and Table~\ref{tab:tesla-ood-cifar} validate the effectiveness of OE~\cite{hendrycks2018deep} in limited-data scenarios, \ie DD settings. For instance, TrustDD (with OE) increases the AUROC metric by $\sim 35.80\%$ when InD and OOD datasets are CIFAR100 and gaussian noise, respectively. Table~\ref{tab:dsa-ood-cifar} and Table~\ref{tab:tesla-ood-cifar} illustrate consistent improvements of TrustDD (with OE) over DD.
So the answer is yes. TrustDD could not only distill knowledge from InD samples, but also from OOD data. Figure~\ref{subfig:rand-distill} further demonstrates that the distilled outliers could be informative.

\vspace{-13pt}
\paragraph{POE even surpasses the SOTA OE.}
The main advantage of POE is that it does not require real outliers. With fewer efforts to carefully collect outlier data compared with OE, POE even surpasses OE for OOD detection as shown in Table~\ref{tab:dsa-ood-cifar} and Table~\ref{tab:tesla-ood-cifar}. POE significantly improves models' ability to detect OOD samples, especially for random noises. A consistent improvement could be witnessed from OE to POE across various test datasets. Considering models trained on CIFAR100 with TESLA to detect uniform noises, compared to OE, POE shows $\sim44.00\%$ increase in AUROC and $\sim21.36\%$ gains averaged on all 7 test OOD datasets. Thus, POE could serve as a strong baseline for future research.

\vspace{-13pt}
\paragraph{Why does POE work?}
Despite not requiring explicit and curated outlier datasets, POE still outperforms OE in many cases as in Table~\ref{tab:dsa-ood-cifar} and Table~\ref{tab:tesla-ood-cifar}. We argue that \emph{pseudo-outliers} are near-ood samples, with appropriate semantic distance to efficiently enhance OOD performance. Once learned to reject such difficult outliers, models are expected to generalize well to detect various OOD samples across the open space~\cite{lee2018training}. 
Furthermore, POE not only works in the DD setting, but also performs admirably in full dataset scenarios, \ie, traditional OOD settings. Please refer to the Appendix for more details.

\begin{figure}[t]
    \centering
    \includegraphics[width=\linewidth]{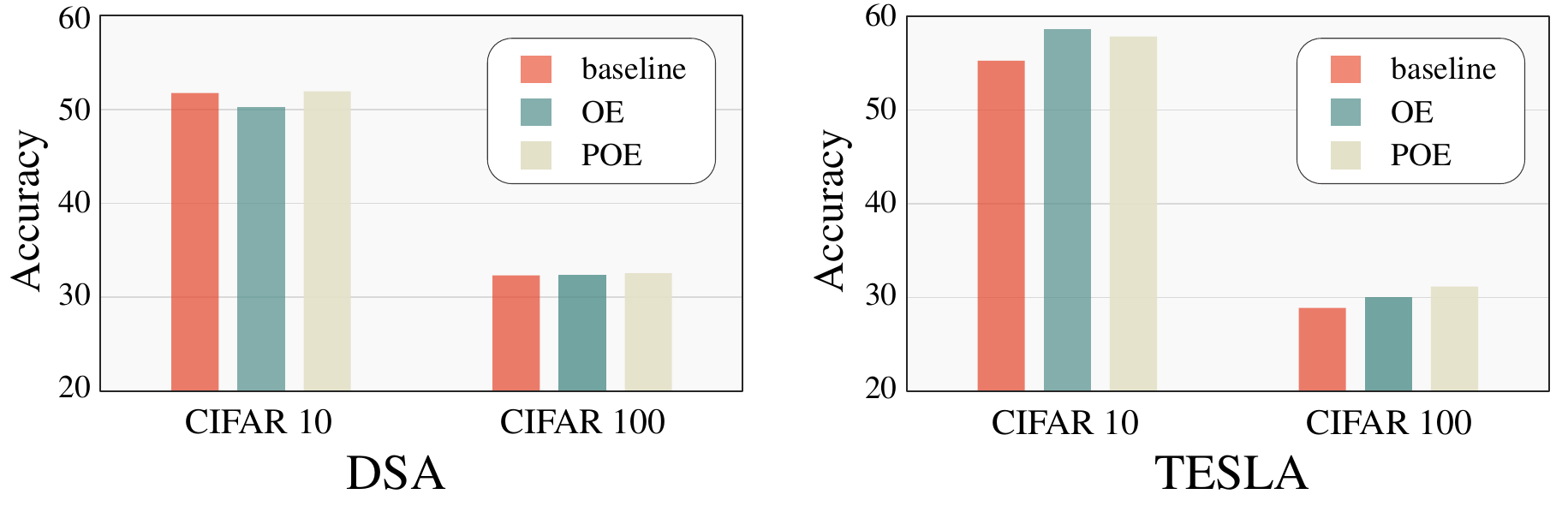}
    \vspace{-20pt}
    \caption{In-distribution accuracy ($\%$) of model trained on distilled data on CIFAR10 and CIFAR100.}
    \label{fig:id-acc}
    \vspace{-10pt}
\end{figure}

Besides, we also evaluate InD performance on CIFAR as in Figure~\ref{fig:id-acc}. TrustDD with both OE and POE boosts OOD performance without the loss of InD classification.

\vspace{-12pt}
\paragraph{Digit Datasets and ImageNet Subsets.}
For digit datasets: MNIST~\cite{lecun1998gradient} and SVHN~\cite{37648}, we implement TrustDD on DSA with IPC=10. Results are shown in Table~\ref{tab:digits}. In this setting, the \emph{pseudo-outliers} contain very narrow semantic information compared with natural image scenarios (\eg, POE in CIFAR) due to the limited semantic range of InD dataset, while OE employs a wide range of outlier semantics, leading to better OOD performance.

\vspace{-5pt}
\begin{table}[!h]
\setlength\tabcolsep{2pt}
\centering
\renewcommand{\arraystretch}{1}
%\scriptsize
\large
\caption{InD classification and OOD detection performance on digit datasets: MNIST and SVHN, with IPC=10.}
\label{tab:digits}
\vspace{-7pt}
\resizebox{.98\linewidth}{!}{
\begin{tabular}{@{}cccccccccc@{}}
\toprule
\multirow{2}{*}{$\mathcal{D}_\textrm{in}$} & \multicolumn{3}{c}{AUROC} & \multicolumn{3}{c}{AUPR-IN} & \multicolumn{3}{c}{Accuracy} \\ \cmidrule(l){2-4} \cmidrule(l){5-7} \cmidrule(l){8-10} 
 & baseline & OE & POE (ours) & baseline & OE & POE (ours) & baseline & OE & POE (ours) \\ \midrule
MNIST & 95.70 & \textbf{99.71} & 99.25 & 96.77 & \textbf{99.75} & 99.45 & 97.91 & \textbf{98.11} & 98.08 \\
SVHN & 84.13 & \textbf{96.75} & 95.52 & 93.19 & \textbf{98.57} & 97.97 & \textbf{79.52} & 78.04 & 79.24 \\ \bottomrule
\end{tabular}
}
\vspace{-5pt}
\end{table}

For higher resolution datasets, we carry out experiments on ImageNet~\cite{5206848}, due to the huge memory-consumption issue of DD, similar to~\cite{cazenavette2022distillation}, we mainly implement TrustDD with IPC=1 on ImageNet Subsets including ImageNette~\cite{imagenette} and ImageFruit~\cite{cazenavette2022distillation}, we also randomly sample another subset, and name it ImageMisc, results are shown in Table~\ref{tab:imagenet}. Both OE and POE improve OOD detection by a large margin. POE outperforms OE in ImageNette and ImageFruit but underperforms OE in ImageMisc, although in all 3 subsets, the performance gap is negligible. When $|\mathcal{S}_\textrm{out}|$ is small, the advantage of POE over OE is relatively subtle, more details are discussed in Section.~\ref{subsec:ablation}.

\vspace{-5pt}
\begin{table}[!h]
\setlength\tabcolsep{2pt}
\centering
\renewcommand{\arraystretch}{1}
\large
\caption{InD classification and OOD detection performance on ImageNet Subsets, with IPC=1.}
\label{tab:imagenet}
\vspace{-7pt}
\resizebox{.98\linewidth}{!}{
\begin{tabular}{@{}cccccccccc@{}}
\toprule
\multirow{2}{*}{$\mathcal{D}_\textrm{in}$} & \multicolumn{3}{c}{AUROC} & \multicolumn{3}{c}{AUPR-IN} & \multicolumn{3}{c}{Accuracy} \\ \cmidrule(l){2-4} \cmidrule(l){5-7} \cmidrule(l){8-10}
 & baseline & OE & POE (ours) & baseline & OE & POE (ours) & baseline & OE & POE (ours) \\ \midrule
ImageNette & 59.52 & 65.35 & \textbf{67.23} & 73.68 & 79.49 & \textbf{80.09} & 47.88 & \textbf{57.00} & 55.72 \\
ImageFruit & 51.18 & 56.90 & \textbf{57.14} & 67.37 & 72.05 & \textbf{72.21} & 28.16 & 30.96 & \textbf{32.16} \\
ImageMisc & 58.88 & \textbf{64.44} & 64.01 & 75.52 & \textbf{79.36} & 78.66 & 43.88 & \textbf{48.68} & 47.12 \\ \bottomrule
\end{tabular}
}
\vspace{-5pt}
\end{table}

\vspace{-12pt}
\paragraph{TrustDD and other OOD scores are complementary.}
To test the generalization of TrustDD across different OOD scores, we also evaluate two well-known detection scores: Maximum Logit Score (MLS)~\cite{vaze2022openset} and Energy Score (Energy)~\cite{liu2020energy} in addition to MSP~\cite{hendrycks2017a}. Here we use DSA with IPC=10, results in Table~\ref{tab:ood-score} show consistent improvements in TrustDD. Furthermore, switching the score from MSP to Energy results in a greater improvement under the POE condition ($\sim 8.95\%$) compared to baseline ($\sim 1.10\%$) and OE ($\sim 1.70\%$). TrustDD (with POE) and score functions are from different perspectives, combining TrustDD with suitable scores could enhance OOD performance by a larger margin than ordinary DD and OE.

\vspace{-7pt}
\begin{table}[!h]
\setlength\tabcolsep{2pt}
\centering
\renewcommand{\arraystretch}{0.9}
%\scriptsize
\large
\caption{OOD detection performance on various scores. Mean AUROC is reported.}
\label{tab:ood-score}
\vspace{-10pt}
\resizebox{.98\linewidth}{!}{
\begin{tabular}{@{}cccccccccc@{}}
\toprule
\multirow{2}{*}{dataset} & \multicolumn{3}{c}{MSP} & \multicolumn{3}{c}{MLS} & \multicolumn{3}{c}{Energy} \\ \cmidrule(l){2-4} \cmidrule(l){5-7} \cmidrule(l){8-10} 
 & baseline & OE & POE(ours) & baseline & OE & POE(ours) & baseline & OE & POE(ours) \\ \midrule
CIFAR10 & 64.06 & 71.30 & \textbf{72.50} & 65.39 & 73.00 & \textbf{78.32} & 65.16 & 73.00 & \textbf{81.45} \\
CIFAR100 & 58.53 & 66.85 & \textbf{72.66} & 60.58 & 70.30 & \textbf{76.44} & 60.65 & 71.52 & \textbf{77.74} \\ \bottomrule
\end{tabular}
}
\vspace{-7pt}
\end{table}

\vspace{-5pt}
\paragraph{Cross-Architecture Generalization.}
TrustDD is from the data perspective, here we validate its generalization abilities across different backbones. We implement DSA on CIFAR10 with IPC=50. Images are distilled on ConvNet, and used to train ConvNet, AlexNet~\cite{krizhevsky2017imagenet}, VGG-11~\cite{vgg} and ResNet-18~\cite{He_2016_CVPR} to test InD and OOD performance. Results in Table~\ref{tab:cross-architecture} show TrustDD could generalize well to unseen model architectures, and POE outperforms OE consistently. Interestingly, when transferring to unseen backbones, InD classification degrades all the time, while OOD performance for OE and POE even increases. 

\vspace{-5pt}
\begin{table}[!h]
\setlength\tabcolsep{4pt}
\centering
\renewcommand{\arraystretch}{1}
\large
\caption{Test performance on various network architectures. Results are shown in the form of \texttt{baseline}/\texttt{OE}/\texttt{POE}.}
\label{tab:cross-architecture}
\vspace{-7pt}
\resizebox{\linewidth}{!}{
\begin{tabular}{@{}ccccc@{}}
\toprule
Metric & ConvNet & AlexNet & VGG & ResNet \\ \midrule
AUROC & 66.08/73.74/\textbf{76.75} & 61.05/79.82/\textbf{89.54} & 58.37/79.30/\textbf{88.65} & 61.30/77.08/\textbf{81.13} \\
AUPR-IN & 71.24/77.51/\textbf{80.16} & 64.99/81.18/\textbf{90.05} & 62.61/80.37/\textbf{89.08} & 65.98/77.85/\textbf{80.52} \\
Accuracy & \textbf{60.55}/59.01/60.22 & 56.23/56.88/\textbf{58.36} & 55.02/\textbf{58.08}/56.29 & 50.00/50.44/\textbf{51.25} \\ \bottomrule
\end{tabular}
}
\vspace{-5pt}
\end{table}

\subsection{Further Analysis} \label{subsec:further-analysis}

\begin{figure}
    \centering
    \begin{subfigure}{.48\linewidth}
        \includegraphics[width=\linewidth]{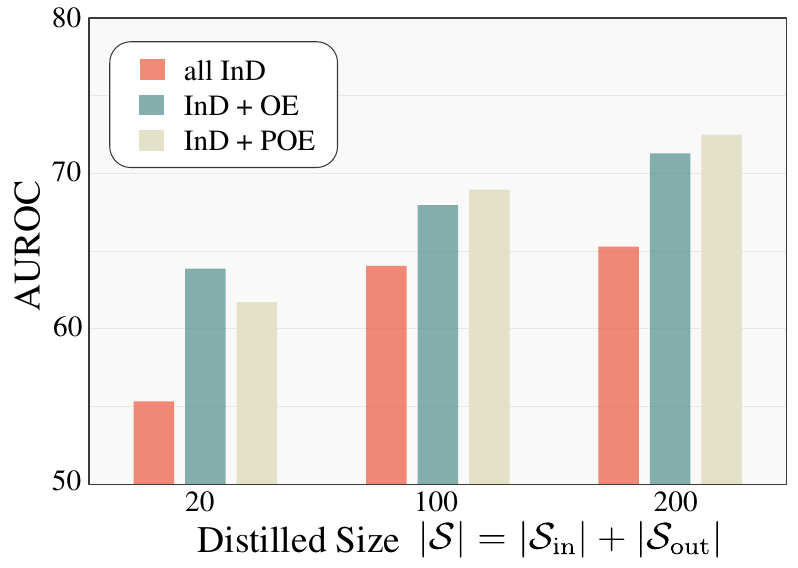}
        \vspace{-18pt}
        \caption{\scriptsize The necessity of distilling $\mathcal{S}_\textrm{out}$.}
        \label{subfig:why-not-all-id}
    \end{subfigure}
    \hfill
    \begin{subfigure}{.48\linewidth}
        \includegraphics[width=\linewidth]{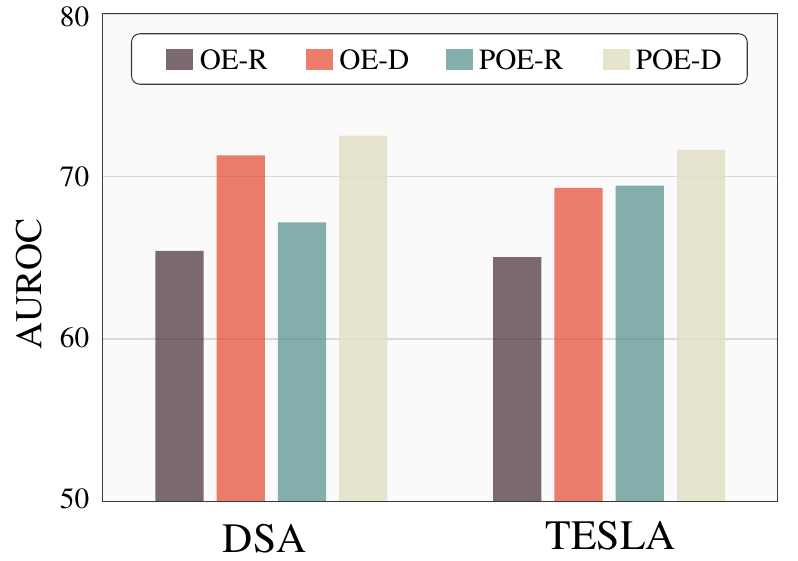}
        \vspace{-18pt}
        \caption{\scriptsize Distilled $\mathcal{S}_\textrm{out}$ is informative.}
        \label{subfig:rand-distill}
    \end{subfigure}
    \vspace{-10pt}
    \caption{The rationale of TrustDD. (a). OOD detection performance of distilling only InD and distilling both InD and OOD to same size $|\mathcal{S}|$. All InD: $|\mathcal{S}_\textrm{in}|=|\mathcal{S}|,|\mathcal{S}_\textrm{out}|=0$. InD+OE and InD+POE: $|\mathcal{S}_\textrm{in}|=|\mathcal{S}_\textrm{out}|=\frac{1}{2}|\mathcal{S}|$. (b). OOD detection performance of OE-R, OE-D, POE-R and POE-D, where `-R' denotes `randomly selected' outliers while `-D' denotes `distilled' outliers via TrustDD.}
    \label{fig:ablation}
    \vspace{-10pt}
\end{figure}

\vspace{-3pt}
\paragraph{Why not simply distill more InD data?}
Vaze \etal~\cite{vaze2022openset} found a positive correlation between ID and OOD performance. It is natural to consider simply increasing the size of $\mathcal{S}_\textrm{in}$ (\ie, without explicitly distilling outliers like TrustDD) to train better classifiers, which are also better OOD detectors, however, Figure~\ref{subfig:why-not-all-id} shows that it is ineffective to improve OOD detection by simply distill more InD images. When raising $|\mathcal{S}|$ from 100 to 200, AUROC increases only $\sim1.23\%$, but $\sim3.54\%$ for TrustDD on our proposed POE.

\vspace{-12pt}
\paragraph{Distilled outliers and random outliers.}
TrustDD not only distills informative InD samples, but also informative outliers. By replacing the distilled outliers with randomly selected outliers, the performance degrades significantly. Figure~\ref{subfig:rand-distill} validates the effectiveness of outlier distillation.

% \begin{figure}
%     \centering
%     \includegraphics[width=.8\linewidth]{figs/why-not-all-id.pdf}
%     \vspace{-10pt}
%     \caption{OOD detection performance of distilling only InD and distilling both InD and OOD to same size $|\mathcal{S}|$. All InD: $|\mathcal{S}_\textrm{in}|=|\mathcal{S}|,|\mathcal{S}_\textrm{out}|=0$. InD+OE and InD+POE: $|\mathcal{S}_\textrm{in}|=|\mathcal{S}_\textrm{out}|=\frac{1}{2}|\mathcal{S}|$.}
%     \label{fig:why-not-all-id}
%     \vspace{-5pt}
% \end{figure}

% \begin{figure}
% \vspace{-5pt}
%     \centering
%     \includegraphics[width=.8\linewidth]{figs/rand-distill.pdf}
%     \vspace{-10pt}
%     \caption{OOD detection performance of OE-R, OE-D, POE-R and POE-D, where `-R' denotes `randomly selected' outliers while `-D' denotes `distilled' outliers via TrustDD.}
%     \label{fig:rand-distill}
%     \vspace{-10pt}
% \end{figure}

\vspace{-12pt}
\paragraph{Why not distill both InD and outliers into InD?}
There is a more efficient way by distilling both $\mathcal{T}_\textrm{in}$ and $\mathcal{T}_\textrm{out}$ together into a single set $\mathcal{S}_\textrm{single}$ and make models trained on $\mathcal{S}_\textrm{single}$ competent in both InD and OOD tasks, \ie, leave out the second term in Eq.~\eqref{eq:loss-on-distilled-set} and make $\mathcal{S}_\textrm{single}=\mathcal{S}$. Nevertheless, distilling both $\mathcal{T}_\textrm{in}$ and $\mathcal{T}_\textrm{out}$ together into $\mathcal{S}_\textrm{single}$ will decrease the information content of $\mathcal{T}_\textrm{in}$ in $\mathcal{S}$ to save room for $\mathcal{T}_\textrm{out}$, leading to degraded classification performance, which could in turn result in poor OOD performance~\cite{vaze2022openset}, even worse than baseline. More details are shown in Appendix.

\subsection{Ablation Studies} \label{subsec:ablation}
\vspace{-3pt}
\paragraph{Different IPCs (Image Per Class).}
To validate the effectiveness of TrustDD on different distillation size $|\mathcal{S}|$, we further conduct TrustDD experiments on CIFAR10 with IPC=1/10/50. Results in Table~\ref{tab:ipc} show TrustDD consistently boosts OOD performance without the degradation of InD classification. In Table~\ref{tab:ipc}, as IPC grows, the advantage of POE over OE becomes increasingly obvious. An intuitive explanation is that POE relies on \emph{pseudo-outliers} (\ie, near-ood samples) to model boundaries between InD and OOD, as $|\mathcal{S}_\textrm{out}|$ grows, such boundaries are increasingly accurate, leading to superior OOD performance of POE.

\vspace{-7pt}
\begin{table}[!h]
\setlength\tabcolsep{2pt}
\centering
\renewcommand{\arraystretch}{0.9}
\scriptsize
\caption{InD and OOD performance on different IPCs.}
\label{tab:ipc}
\vspace{-10pt}
\resizebox{.9\linewidth}{!}{
\begin{tabular}{@{}ccccccc@{}}
\toprule
\multirow{2}{*}{IPC} & \multicolumn{3}{c}{AUROC} & \multicolumn{3}{c}{Accuracy} \\ \cmidrule(l){2-4} \cmidrule(l){5-7}  
 & baseline & OE & POE (ours) & baseline & OE & POE (ours) \\ \midrule
1 & 54.91 & \textbf{63.89} & 61.73 & 29.07 & 31.96 & \textbf{32.15} \\
10 & 64.06 & 71.30 & \textbf{72.50} & 51.74 & 50.24 & \textbf{51.91} \\
50 & 66.08 & 73.74 & \textbf{76.75} & \textbf{60.55} & 59.01 & 60.22 \\ \bottomrule
\end{tabular}
}
\vspace{-10pt}
\end{table}

\begin{figure}
    \centering
    \includegraphics[width=.9\linewidth]{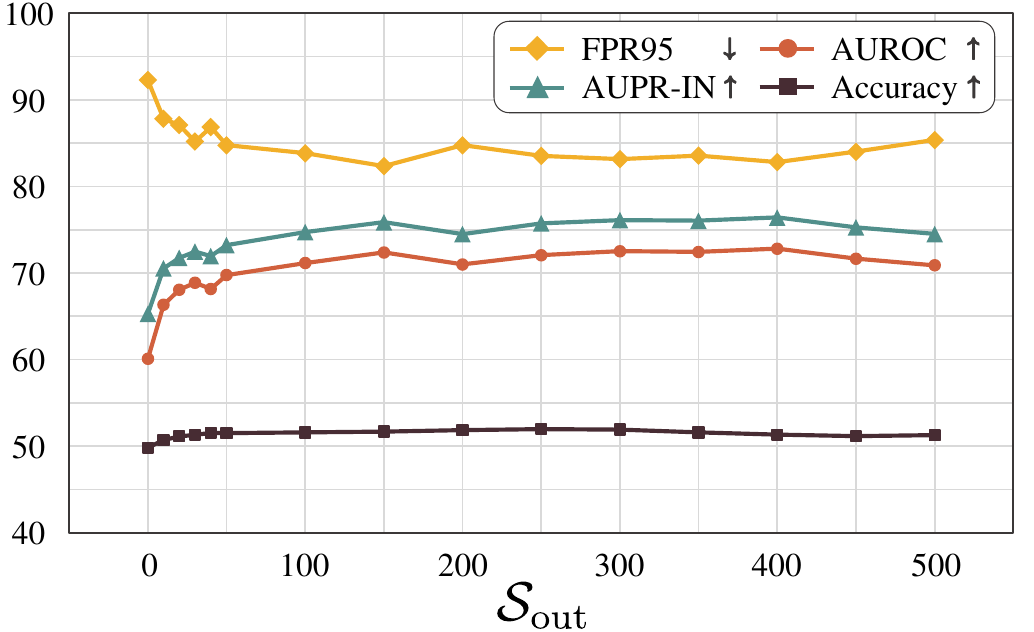}
    \vspace{-10pt}
    \caption{Influence of distilled outlier size. Here, we keep $|\mathcal{S}_\textrm{in}|$ equals to 100, and changes $|\mathcal{S}_\textrm{out}|$.}
    \label{fig:oe-num}
    \vspace{-5pt}
\end{figure}

\vspace{-13pt}
\paragraph{The Influence of $|\mathcal{S}_\textrm{out}|$.}
As described in Section~\ref{subsec:exp-setup}, TrustDD additionally condenses outlier data $\mathcal{S}_\textrm{out}$ to train better OOD detectors at the cost of slightly larger distilled size $|\mathcal{S}|$, to balance efficiency and trustworthiness, we keep InD IPC=10, \ie $|\mathcal{S}_\textrm{in}|=100$, and distill various numbers of outliers, the results are shown in Figure~\ref{fig:oe-num}. When $|\mathcal{S}_\textrm{in}|=100$, it is appropriate to set $|\mathcal{S}_\textrm{out}|\in[50,150]$ according to the practical issues.

\subsection{Visualizations} \label{subsec:visualization}
\begin{figure}[!h]
    \centering
    \begin{subfigure}{0.45\linewidth}
     \centering
     \includegraphics[width=\linewidth]{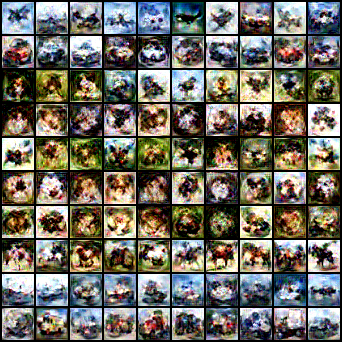}
     \vspace{-18pt}
     \subcaption*{Distilled InD $\mathcal{S}_\textrm{in}$}
    \end{subfigure}
    \hfill
    \begin{subfigure}{0.45\linewidth}
     \centering
     \includegraphics[width=\linewidth]{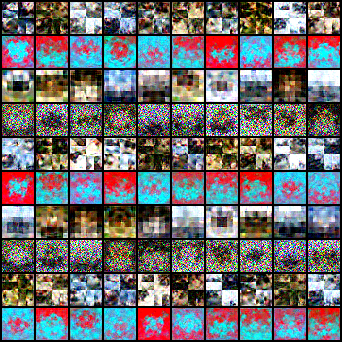}
     \vspace{-18pt}
     \subcaption*{Distilled OOD $\mathcal{S}_\textrm{out}$}
    \end{subfigure}
    \vspace{-12pt}
    \caption{Visualization of TrustDD distill images on CIFAR10~\cite{krizhevsky2009learning} with IPC=10.}
    \label{fig:cifar-vis}
    \vspace{-5pt}
\end{figure}

\begin{figure}
    \centering
    \begin{subfigure}{\linewidth}
     \centering
     \includegraphics[width=\linewidth]{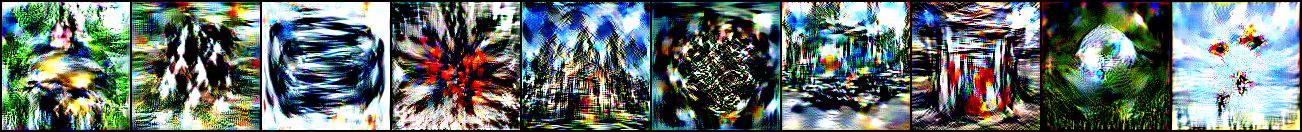}
     \vspace{-18pt}
     \subcaption*{Distilled InD $\mathcal{S}_\textrm{in}$}
    \end{subfigure}
    \vfill
    \begin{subfigure}{\linewidth}
     \centering
     \includegraphics[width=\linewidth]{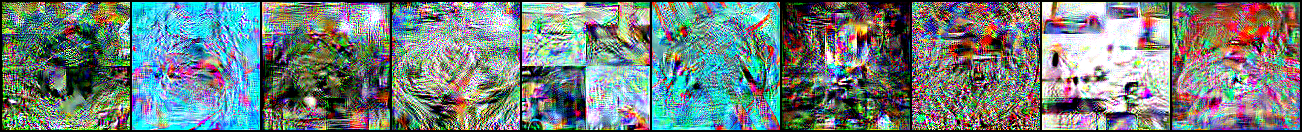}
     \vspace{-18pt}
     \subcaption*{Distilled OOD $\mathcal{S}_\textrm{out}$}
    \end{subfigure}
    \vspace{-10pt}
    \caption{Visualization of TrustDD distill images on ImageNette~\cite{imagenette} with IPC=1.}
    \label{fig:imagenette-vis}
    \vspace{-10pt}
\end{figure}

\vspace{-3pt}
Distilled images are visualized in Figure~\ref{fig:cifar-vis} and Figure~\ref{fig:imagenette-vis}. When distilling $\mathcal{S}_\textrm{out}$, we explicitly make each $\boldsymbol{s}^i_\textrm{out}$ align with one corruption. In Figure~\ref{fig:cifar-vis}, the first four rows in $\mathcal{S}_\textrm{out}$ correspond to \texttt{jigsaw}, \texttt{invert}, \texttt{mosaic} and \texttt{speckle} respectively, similar results could be observed in Figure~\ref{fig:imagenette-vis}. Please refer to the Appendix for more visualizations.

\section{Conclusion}

In this paper, we propose a novel learning paradigm called \emph{Trustworthy Dataset Distillation} (TrustDD) from the data perspective, which is the first attempt to take both efficiency and trustworthiness into consideration for dataset distillation. TrustDD explicitly condenses in-distribution samples and outliers into separate tiny yet informative sets, upon which models trained perform competently on InD classification and OOD detection simultaneously. We further introduce to generate \emph{pseudo-outliers} via InD corruption, and the proposed POE achieves superior performance compared with SOTA method OE even without the accessibility to real outlier data. Without degradation in InD classification performance, TrustDD with POE makes DD more trustworthy and more applicable to the open-world, and serves as a strong baseline for future research in the under-explored open-world setting in DD.

%\newpage
{\small
\bibliographystyle{ieee_fullname}
\bibliography{egbib}

\begin{thebibliography}{10}\itemsep=-1pt

\bibitem{aljundi2019gradient}
Rahaf Aljundi, Min Lin, Baptiste Goujaud, and Yoshua Bengio.
\newblock Gradient based sample selection for online continual learning.
\newblock {\em Advances in neural information processing systems}, 32, 2019.

\bibitem{bogdoll2022anomaly}
Daniel Bogdoll, Maximilian Nitsche, and J~Marius Z{\"o}llner.
\newblock Anomaly detection in autonomous driving: A survey.
\newblock In {\em Proceedings of the IEEE/CVF conference on computer vision and
  pattern recognition}, pages 4488--4499, 2022.

\bibitem{notmnist}
Yaroslav Bulatov.
\newblock Notmnist dataset.
\newblock {\em Google (Books/OCR), Tech. Rep.[Online]. Available:
  http://yaroslavvb. blogspot. it/2011/09/notmnist-dataset. html}, 2, 2011.

\bibitem{cazenavette2022distillation}
George Cazenavette, Tongzhou Wang, Antonio Torralba, Alexei~A. Efros, and
  Jun-Yan Zhu.
\newblock Dataset distillation by matching training trajectories.
\newblock In {\em Proceedings of the IEEE/CVF Conference on Computer Vision and
  Pattern Recognition}, 2022.

\bibitem{cimpoi2014describing}
Mircea Cimpoi, Subhransu Maji, Iasonas Kokkinos, Sammy Mohamed, and Andrea
  Vedaldi.
\newblock Describing textures in the wild.
\newblock In {\em Proceedings of the IEEE conference on computer vision and
  pattern recognition}, pages 3606--3613, 2014.

\bibitem{tiny}
cs231n.stanford.edu.
\newblock Cs231n: Convolutional neural networks for visual recognition.

\bibitem{cui2022scaling}
Justin Cui, Ruochen Wang, Si Si, and Cho-Jui Hsieh.
\newblock Scaling up dataset distillation to imagenet-1k with constant memory.
\newblock {\em arXiv preprint arXiv:2211.10586}, 2022.

\bibitem{5206848}
Jia Deng, Wei Dong, Richard Socher, Li-Jia Li, Kai Li, and Li Fei-Fei.
\newblock Imagenet: A large-scale hierarchical image database.
\newblock In {\em 2009 IEEE Conference on Computer Vision and Pattern
  Recognition}, pages 248--255, 2009.

\bibitem{pmlr-v162-dong22c}
Tian Dong, Bo Zhao, and Lingjuan Lyu.
\newblock Privacy for free: How does dataset condensation help privacy?
\newblock In Kamalika Chaudhuri, Stefanie Jegelka, Le Song, Csaba Szepesvari,
  Gang Niu, and Sivan Sabato, editors, {\em Proceedings of the 39th
  International Conference on Machine Learning}, volume 162 of {\em Proceedings
  of Machine Learning Research}, pages 5378--5396. PMLR, 17--23 Jul 2022.

\bibitem{imagenette}
Fastai.
\newblock Fastai/imagenette: A smaller subset of 10 easily classified classes
  from imagenet, and a little more french.

\bibitem{He_2016_CVPR}
Kaiming He, Xiangyu Zhang, Shaoqing Ren, and Jian Sun.
\newblock Deep residual learning for image recognition.
\newblock In {\em Proceedings of the IEEE Conference on Computer Vision and
  Pattern Recognition (CVPR)}, June 2016.

\bibitem{pmlr-v162-hendrycks22a}
Dan Hendrycks, Steven Basart, Mantas Mazeika, Andy Zou, Joseph Kwon,
  Mohammadreza Mostajabi, Jacob Steinhardt, and Dawn Song.
\newblock Scaling out-of-distribution detection for real-world settings.
\newblock In Kamalika Chaudhuri, Stefanie Jegelka, Le Song, Csaba Szepesvari,
  Gang Niu, and Sivan Sabato, editors, {\em Proceedings of the 39th
  International Conference on Machine Learning}, volume 162 of {\em Proceedings
  of Machine Learning Research}, pages 8759--8773. PMLR, 17--23 Jul 2022.

\bibitem{hendrycks2017a}
Dan Hendrycks and Kevin Gimpel.
\newblock A baseline for detecting misclassified and out-of-distribution
  examples in neural networks.
\newblock In {\em International Conference on Learning Representations}, 2017.

\bibitem{hendrycks2018deep}
Dan Hendrycks, Mantas Mazeika, and Thomas Dietterich.
\newblock Deep anomaly detection with outlier exposure.
\newblock In {\em International Conference on Learning Representations}, 2019.

\bibitem{hendrycks2019using}
Dan Hendrycks, Mantas Mazeika, Saurav Kadavath, and Dawn Song.
\newblock Using self-supervised learning can improve model robustness and
  uncertainty.
\newblock {\em Advances in neural information processing systems}, 32, 2019.

\bibitem{hendrycks2021natural}
Dan Hendrycks, Kevin Zhao, Steven Basart, Jacob Steinhardt, and Dawn Song.
\newblock Natural adversarial examples.
\newblock In {\em Proceedings of the IEEE/CVF Conference on Computer Vision and
  Pattern Recognition}, pages 15262--15271, 2021.

\bibitem{hinton2015distilling}
Geoffrey Hinton, Oriol Vinyals, and Jeff Dean.
\newblock Distilling the knowledge in a neural network.
\newblock {\em arXiv preprint arXiv:1503.02531}, 2015.

\bibitem{howard2017mobilenets}
Andrew~G Howard, Menglong Zhu, Bo Chen, Dmitry Kalenichenko, Weijun Wang,
  Tobias Weyand, Marco Andreetto, and Hartwig Adam.
\newblock Mobilenets: Efficient convolutional neural networks for mobile vision
  applications.
\newblock {\em arXiv preprint arXiv:1704.04861}, 2017.

\bibitem{9864101}
Hongzhi Huang, Yu Wang, Qinghua Hu, and Ming-Ming Cheng.
\newblock Class-specific semantic reconstruction for open set recognition.
\newblock {\em IEEE Transactions on Pattern Analysis and Machine Intelligence},
  pages 1--14, 2022.

\bibitem{jacob2018quantization}
Benoit Jacob, Skirmantas Kligys, Bo Chen, Menglong Zhu, Matthew Tang, Andrew
  Howard, Hartwig Adam, and Dmitry Kalenichenko.
\newblock Quantization and training of neural networks for efficient
  integer-arithmetic-only inference.
\newblock In {\em Proceedings of the IEEE conference on computer vision and
  pattern recognition}, pages 2704--2713, 2018.

\bibitem{krizhevsky2009learning}
Alex Krizhevsky, Geoffrey Hinton, et~al.
\newblock Learning multiple layers of features from tiny images.
\newblock 2009.

\bibitem{krizhevsky2017imagenet}
Alex Krizhevsky, Ilya Sutskever, and Geoffrey~E Hinton.
\newblock Imagenet classification with deep convolutional neural networks.
\newblock {\em Communications of the ACM}, 60(6):84--90, 2017.

\bibitem{lecun1998gradient}
Yann LeCun, L{\'e}on Bottou, Yoshua Bengio, and Patrick Haffner.
\newblock Gradient-based learning applied to document recognition.
\newblock {\em Proceedings of the IEEE}, 86(11):2278--2324, 1998.

\bibitem{lee2018training}
Kimin Lee, Honglak Lee, Kibok Lee, and Jinwoo Shin.
\newblock Training confidence-calibrated classifiers for detecting
  out-of-distribution samples.
\newblock In {\em International Conference on Learning Representations}, 2018.

\bibitem{lee2018simple}
Kimin Lee, Kibok Lee, Honglak Lee, and Jinwoo Shin.
\newblock A simple unified framework for detecting out-of-distribution samples
  and adversarial attacks.
\newblock {\em Advances in neural information processing systems}, 31, 2018.

\bibitem{pmlr-v162-lee22b}
Saehyung Lee, Sanghyuk Chun, Sangwon Jung, Sangdoo Yun, and Sungroh Yoon.
\newblock Dataset condensation with contrastive signals.
\newblock In Kamalika Chaudhuri, Stefanie Jegelka, Le Song, Csaba Szepesvari,
  Gang Niu, and Sivan Sabato, editors, {\em Proceedings of the 39th
  International Conference on Machine Learning}, volume 162 of {\em Proceedings
  of Machine Learning Research}, pages 12352--12364. PMLR, 17--23 Jul 2022.

\bibitem{liang2018enhancing}
Shiyu Liang, Yixuan Li, and R. Srikant.
\newblock Enhancing the reliability of out-of-distribution image detection in
  neural networks.
\newblock In {\em International Conference on Learning Representations}, 2018.

\bibitem{liu2020energy}
Weitang Liu, Xiaoyun Wang, John Owens, and Yixuan Li.
\newblock Energy-based out-of-distribution detection.
\newblock {\em Advances in neural information processing systems},
  33:21464--21475, 2020.

\bibitem{loo2022efficient}
Noel Loo, Ramin Hasani, Alexander Amini, and Daniela Rus.
\newblock Efficient dataset distillation using random feature approximation.
\newblock In Alice~H. Oh, Alekh Agarwal, Danielle Belgrave, and Kyunghyun Cho,
  editors, {\em Advances in Neural Information Processing Systems}, 2022.

\bibitem{menghani2021efficient}
Gaurav Menghani.
\newblock Efficient deep learning: A survey on making deep learning models
  smaller, faster, and better.
\newblock {\em ACM Computing Surveys}, 2021.

\bibitem{37648}
Yuval Netzer, Tao Wang, Adam Coates, Alessandro Bissacco, Bo Wu, and Andrew~Y.
  Ng.
\newblock Reading digits in natural images with unsupervised feature learning.
\newblock In {\em NIPS Workshop on Deep Learning and Unsupervised Feature
  Learning 2011}, 2011.

\bibitem{nguyen2021dataset}
Timothy Nguyen, Zhourong Chen, and Jaehoon Lee.
\newblock Dataset meta-learning from kernel ridge-regression.
\newblock In {\em International Conference on Learning Representations}, 2021.

\bibitem{NEURIPS2021_299a23a2}
Timothy Nguyen, Roman Novak, Lechao Xiao, and Jaehoon Lee.
\newblock Dataset distillation with infinitely wide convolutional networks.
\newblock In M. Ranzato, A. Beygelzimer, Y. Dauphin, P.S. Liang, and J.~Wortman
  Vaughan, editors, {\em Advances in Neural Information Processing Systems},
  volume~34, pages 5186--5198. Curran Associates, Inc., 2021.

\bibitem{radford2021learning}
Alec Radford, Jong~Wook Kim, Chris Hallacy, Aditya Ramesh, Gabriel Goh,
  Sandhini Agarwal, Girish Sastry, Amanda Askell, Pamela Mishkin, Jack Clark,
  et~al.
\newblock Learning transferable visual models from natural language
  supervision.
\newblock In {\em International conference on machine learning}, pages
  8748--8763. PMLR, 2021.

\bibitem{sachdeva2023data}
Noveen Sachdeva and Julian McAuley.
\newblock Data distillation: A survey.
\newblock {\em arXiv preprint arXiv:2301.04272}, 2023.

\bibitem{sener2018active}
Ozan Sener and Silvio Savarese.
\newblock Active learning for convolutional neural networks: A core-set
  approach.
\newblock In {\em International Conference on Learning Representations}, 2018.

\bibitem{vgg}
K. Simonyan and A. Zisserman.
\newblock Very deep convolutional networks for large-scale image recognition.
\newblock In {\em International Conference on Learning Representations}, May
  2015.

\bibitem{pmlr-v119-such20a}
Felipe~Petroski Such, Aditya Rawal, Joel Lehman, Kenneth Stanley, and Jeffrey
  Clune.
\newblock Generative teaching networks: Accelerating neural architecture search
  by learning to generate synthetic training data.
\newblock In Hal~Daumé III and Aarti Singh, editors, {\em Proceedings of the
  37th International Conference on Machine Learning}, volume 119 of {\em
  Proceedings of Machine Learning Research}, pages 9206--9216. PMLR, 13--18 Jul
  2020.

\bibitem{tack2020csi}
Jihoon Tack, Sangwoo Mo, Jongheon Jeong, and Jinwoo Shin.
\newblock Csi: Novelty detection via contrastive learning on distributionally
  shifted instances.
\newblock {\em Advances in neural information processing systems},
  33:11839--11852, 2020.

\bibitem{van2018inaturalist}
Grant Van~Horn, Oisin Mac~Aodha, Yang Song, Yin Cui, Chen Sun, Alex Shepard,
  Hartwig Adam, Pietro Perona, and Serge Belongie.
\newblock The inaturalist species classification and detection dataset.
\newblock In {\em Proceedings of the IEEE conference on computer vision and
  pattern recognition}, pages 8769--8778, 2018.

\bibitem{vaze2022openset}
Sagar Vaze, Kai Han, Andrea Vedaldi, and Andrew Zisserman.
\newblock Open-set recognition: a good closed-set classifier is all you need?
\newblock In {\em International Conference on Learning Representations}, 2022.

\bibitem{wang2022vim}
Haoqi Wang, Zhizhong Li, Litong Feng, and Wayne Zhang.
\newblock Vim: Out-of-distribution with virtual-logit matching.
\newblock In {\em Proceedings of the IEEE/CVF Conference on Computer Vision and
  Pattern Recognition}, pages 4921--4930, 2022.

\bibitem{wang2022cafe}
Kai Wang, Bo Zhao, Xiangyu Peng, Zheng Zhu, Shuo Yang, Shuo Wang, Guan Huang,
  Hakan Bilen, Xinchao Wang, and Yang You.
\newblock Cafe: Learning to condense dataset by aligning features.
\newblock In {\em Proceedings of the IEEE/CVF Conference on Computer Vision and
  Pattern Recognition}, pages 12196--12205, 2022.

\bibitem{wang2018dataset}
Tongzhou Wang, Jun-Yan Zhu, Antonio Torralba, and Alexei~A Efros.
\newblock Dataset distillation.
\newblock {\em arXiv preprint arXiv:1811.10959}, 2018.

\bibitem{10.1007/978-3-031-16452-1_4}
Julia Wolleb, Florentin Bieder, Robin Sandk{\"u}hler, and Philippe~C. Cattin.
\newblock Diffusion models for medical anomaly detection.
\newblock In Linwei Wang, Qi Dou, P.~Thomas Fletcher, Stefanie Speidel, and
  Shuo Li, editors, {\em Medical Image Computing and Computer Assisted
  Intervention -- MICCAI 2022}, pages 35--45, Cham, 2022. Springer Nature
  Switzerland.

\bibitem{xiao2017/online}
Han Xiao, Kashif Rasul, and Roland Vollgraf.
\newblock Fashion-mnist: a novel image dataset for benchmarking machine
  learning algorithms, 2017.

\bibitem{xu2015turkergaze}
Pingmei Xu, Krista~A Ehinger, Yinda Zhang, Adam Finkelstein, Sanjeev~R
  Kulkarni, and Jianxiong Xiao.
\newblock Turkergaze: Crowdsourcing saliency with webcam based eye tracking.
\newblock {\em arXiv preprint arXiv:1504.06755}, 2015.

\bibitem{yang2022openood}
Jingkang Yang, Pengyun Wang, Dejian Zou, Zitang Zhou, Kunyuan Ding, WENXUAN
  PENG, Haoqi Wang, Guangyao Chen, Bo Li, Yiyou Sun, Xuefeng Du, Kaiyang Zhou,
  Wayne Zhang, Dan Hendrycks, Yixuan Li, and Ziwei Liu.
\newblock Open{OOD}: Benchmarking generalized out-of-distribution detection.
\newblock In {\em Thirty-sixth Conference on Neural Information Processing
  Systems Datasets and Benchmarks Track}, 2022.

\bibitem{yu2015lsun}
Fisher Yu, Ari Seff, Yinda Zhang, Shuran Song, Thomas Funkhouser, and Jianxiong
  Xiao.
\newblock Lsun: Construction of a large-scale image dataset using deep learning
  with humans in the loop.
\newblock {\em arXiv preprint arXiv:1506.03365}, 2015.

\bibitem{yu2023dataset}
Ruonan Yu, Songhua Liu, and Xinchao Wang.
\newblock Dataset distillation: A comprehensive review.
\newblock {\em arXiv preprint arXiv:2301.07014}, 2023.

\bibitem{zhao2021DSA}
Bo Zhao and Hakan Bilen.
\newblock Dataset condensation with differentiable siamese augmentation.
\newblock In {\em International Conference on Machine Learning}, 2021.

\bibitem{zhao2023DM}
Bo Zhao and Hakan Bilen.
\newblock Dataset condensation with distribution matching.
\newblock 2023.

\bibitem{zhao2021DC}
Bo Zhao, Konda~Reddy Mopuri, and Hakan Bilen.
\newblock Dataset condensation with gradient matching.
\newblock In {\em International Conference on Learning Representations}, 2021.

\bibitem{zhou2017places}
Bolei Zhou, Agata Lapedriza, Aditya Khosla, Aude Oliva, and Antonio Torralba.
\newblock Places: A 10 million image database for scene recognition.
\newblock {\em IEEE transactions on pattern analysis and machine intelligence},
  40(6):1452--1464, 2017.

\bibitem{zhou2022dataset}
Yongchao Zhou, Ehsan Nezhadarya, and Jimmy Ba.
\newblock Dataset distillation using neural feature regression.
\newblock In {\em Proceedings of the Advances in Neural Information Processing
  Systems (NeurIPS)}, 2022.

\end{thebibliography}
}

\clearpage

\appendix

\section{Appendix} \label{sec:appendix}

\subsection{Datasets}

\subsubsection{InD Datasets}

\paragraph{Digit datasets.}
MNIST~\cite{lecun1998gradient} and SVHN~\cite{37648} are two well-known digit datasets. MNIST is a handwritten digits dataset with $60,000$ training samples and $10,000$ test samples with size $28\times 28$. SVHN is about street view numbers containing $73,257$ and $26,032$ training and test samples with size $32\times 32$, respectively. Both MNIST and SVHN have 10 classes, and MNIST images are grayscale images.

\vspace{-12pt}
\paragraph{CIFAR~\cite{krizhevsky2009learning}.}
CIFAR10 and CIFAR100 are natural color image datasets containing $50,000$ training images and $10,000$ test images, with size $32\times 32$. CIFAR10 has 10 classes while CIFAR100 has 100 classes.

\vspace{-12pt}
\paragraph{ImageNet Subsets.}
Following~\cite{cazenavette2022distillation}, each subset is with 10 classes, and images are down-sampled to $128\times 128$. Here, we mainly use three subsets: ImageNette~\cite{imagenette}, ImageFruit~\cite{cazenavette2022distillation} and ImageMisc. ImageNette and ImageFruit are two pre-existing datasets, while ImageMisc is created in this paper by randomly sampling 10 classes in ImageNet~\cite{5206848} full dataset. Here, we enumerate the specific categories of each subset in Table~\ref{tab:imagenet-subset-class}.

\begin{table}[!h]
\vspace{-5pt}
\centering
\caption{The specific categories of three ImageNet Subsets, which are shown in the form of \texttt{class-id}-\texttt{class-name}.}
\label{tab:imagenet-subset-class}
\vspace{-7pt}
\resizebox{\linewidth}{!}{
\begin{tabular}{@{}cc@{}}
\toprule
Subset & Categories \\ \midrule
ImageNette & \begin{tabular}[c]{@{}c@{}}0-tench, 217-english springer, 482-cassette player, 491-chainsaw, 497-church,\\ 566-horn, 569-dustcart, 571-gas pump, 574-golf ball, 701-chute\end{tabular} \\ \midrule
ImageFruit & \begin{tabular}[c]{@{}c@{}}953-pineapple, 954-banana, 949-strawberry, 950-orange, 951-lemon,\\ 957-pomegranate, 952-fig, 945-bell pepper, 943-cucumber, 948-green apple\end{tabular} \\ \midrule
ImageMisc & \begin{tabular}[c]{@{}c@{}}2-shark, 99-goose, 207-golden retriever, 404-airliner, 430-basketball,\\ 565-freight car, 691-oxygen mask, 851-television, 949-strawberry, 966-red wine\end{tabular} \\ \bottomrule
\end{tabular}
}
\vspace{-15pt}
\end{table}

\vspace{-10pt}
\subsubsection{OOD datasets}
The corresponding test OOD datasets are listed in Table~\ref{tab:id-ood-list}. When the image resolution is not the same, we resize OOD images to the InD image size, \eg, the resized version of Tiny ImageNet~\cite{tiny} and LSUN~\cite{yu2015lsun}.

\subsection{Experimental Details}
\paragraph{Metrics.}
For InD classification, we measure the classification accuracy, For OOD detection, we measure the commonly-used metrics: (1) FPR95: False Positive Rate when True Positive Rate is at $95\%$, which could be understood as the ratio of OOD samples that models misclassify as InD samples when $95\%$ of the in-distribution samples could be detected. (2) AUROC: Area Under the Receiver Operating Characteristic curve. AUROC could be interpreted as the probability that each positive (InD) sample to be assigned a higher score than OOD samples. (3) AUPR-IN and AUPR-OUT: Area under the Precision-Recall curve where in-distribution and out-of-distribution samples are regarded as positives respectively. Both AUROC and AUPR(-IN or -OUT) are threshold-independent metrics.

\vspace{-12pt}
\paragraph{Hyper-parameters.}
The main hyper-parameter is the trade-off weight $\lambda$ of two tasks, we find that the results are not sensitive when $\lambda$ is in $[0.3,0.7]$ through validation experiments (CIFAR serves as the InD dataset and SVHN serves as the validation OOD test dataset), so we simply set $\lambda=0.5$ in all experiments as in~\cite{hendrycks2018deep}. For networks, we mainly use ConvNet, which is with several convolutional layers, each has 128 channels and is equipped with RELU, InstanceNorm and average pooling. The final fully-connected layer outputs the logits.

\begin{figure}[!h]
\vspace{-10pt}
    \centering
    \includegraphics[width=.7\linewidth]{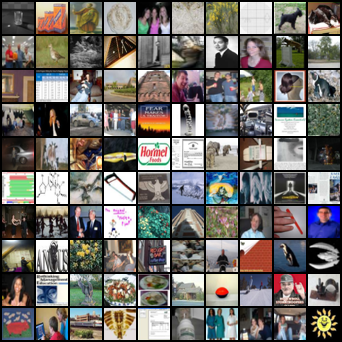}
    \vspace{-7pt}
    \caption{Examples of 300K Random Images in OE.}
    \label{fig:oe_300k}
    \vspace{-10pt}
\end{figure}

\vspace{-12pt}
\paragraph{OE data.}
OE~\cite{hendrycks2018deep} explicitly utilizes curated outliers, in all our experiments, we use 300K Random Images~\cite{hendrycks2018deep}, the example images are shown in Table~\ref{fig:oe_300k}.

\vspace{-12pt}
\paragraph{POE and corruptions.}
Corruptions should noticeably change the semantics of in-distribution samples. For natural scene images, we mainly implement \texttt{jigsaw}, \texttt{invert}, \texttt{mosaic} and \texttt{speckle} as in Section~\ref{subsec:poe}. It is worth noting that \texttt{flip} is a semantic-preserving transformation for natural images, however, this is not the case for digits and characters. Hence, for digits datasets like MNIST~\cite{lecun1998gradient} and SVHN~\cite{37648}, one could also apply \texttt{flip} transformation which slightly shifts the semantics. For digits, $\texttt{flip}$ refers to horizontal (without `0', `1', `8') or vertical (without `0', `1', `3', `8') flips, flipping `0', `1' and `8' does not shift the semantics of digits, so when generating \emph{pseudo-outliers} we simply leave out them. With the ensemble of corruption transformations, one could generate an outlier dataset $\mathcal{T}_\textrm{out}$ from $\mathcal{T}_\textrm{in}$. Ensemble means that we employ various corruptions to different images and collect them together to obtain $\mathcal{T}_\textrm{out}$, here, for each image, we implement one certain corruption.

\begin{figure*}[t]
     \centering
     \begin{subfigure}{0.19\linewidth}
         \centering
         \includegraphics[width=\linewidth]{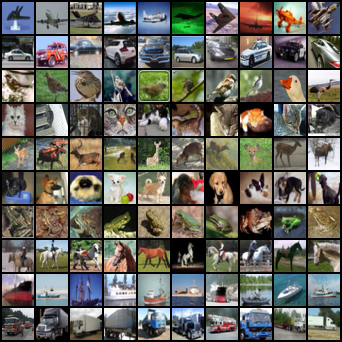}
         \vspace{-15pt}
         \subcaption*{\texttt{original}}
     \end{subfigure}
     \hfill
     \begin{subfigure}{0.19\linewidth}
         \centering
         \includegraphics[width=\linewidth]{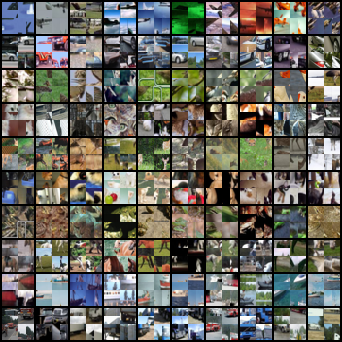}
         \vspace{-15pt}
         \subcaption*{\texttt{jigsaw}}
     \end{subfigure}
     \hfill
     \begin{subfigure}{0.19\linewidth}
         \centering
         \includegraphics[width=\linewidth]{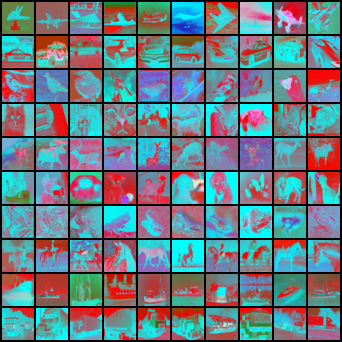}
         \vspace{-15pt}
         \subcaption*{\texttt{invert}}
     \end{subfigure}
     \hfill
     \begin{subfigure}{0.19\linewidth}
         \centering
         \includegraphics[width=\linewidth]{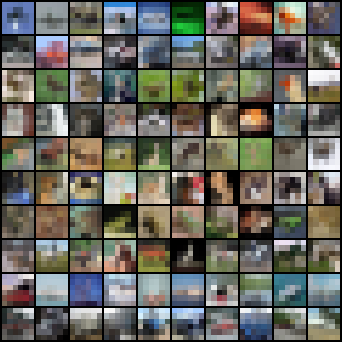}
         \vspace{-15pt}
         \subcaption*{\texttt{mosaic}}
     \end{subfigure}
     \hfill
     \begin{subfigure}{0.19\linewidth}
         \centering
         \includegraphics[width=\linewidth]{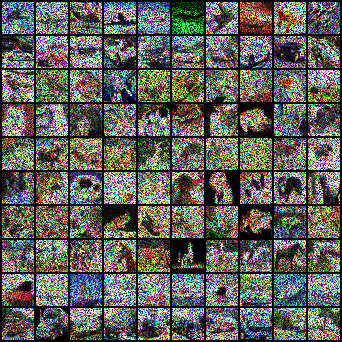}
         \vspace{-15pt}
         \subcaption*{\texttt{speckle}}
     \end{subfigure}
     \vspace{-10pt}
        \caption{Visualization of \emph{pseudo-outliers} on CIFAR10~\cite{krizhevsky2009learning}.}
        \label{fig:cifar-poe-visualization}
    \vspace{-5pt}
\end{figure*}

\begin{figure*}[ht]
     \centering
     \begin{subfigure}{0.19\linewidth}
         \centering
         \includegraphics[width=\linewidth]{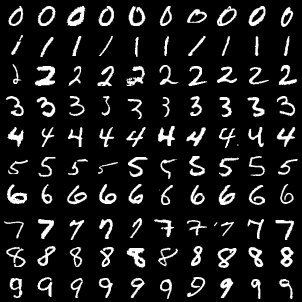}
         \vspace{-15pt}
         \subcaption*{\texttt{original}}
     \end{subfigure}
     \hfill
     \begin{subfigure}{0.19\linewidth}
         \centering
         \includegraphics[width=\linewidth]{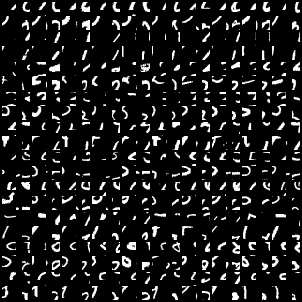}
         \vspace{-15pt}
         \subcaption*{\texttt{jigsaw}}
     \end{subfigure}
     \hfill
     \begin{subfigure}{0.19\linewidth}
         \centering
         \includegraphics[width=\linewidth]{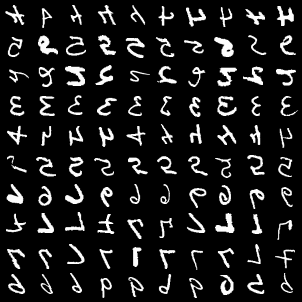}
         \vspace{-15pt}
         \subcaption*{\texttt{flip}}
     \end{subfigure}
     \hfill
     \begin{subfigure}{0.19\linewidth}
         \centering
         \includegraphics[width=\linewidth]{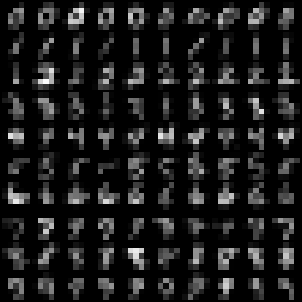}
         \vspace{-15pt}
         \subcaption*{\texttt{mosaic}}
     \end{subfigure}
     \hfill
     \begin{subfigure}{0.19\linewidth}
         \centering
         \includegraphics[width=\linewidth]{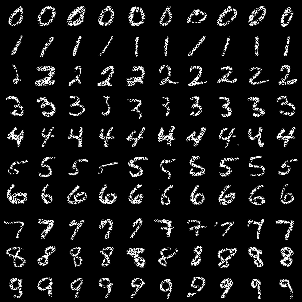}
         \vspace{-15pt}
         \subcaption*{\texttt{speckle}}
     \end{subfigure}
     \vspace{-10pt}
        \caption{Visualization of \emph{pseudo-outliers} on MNIST~\cite{lecun1998gradient}.}
        \label{fig:mnist-poe-visualization}
    \vspace{-5pt}
\end{figure*}

\begin{figure*}[!ht]
     \centering
     \begin{subfigure}{0.19\linewidth}
         \centering
         \includegraphics[width=\linewidth]{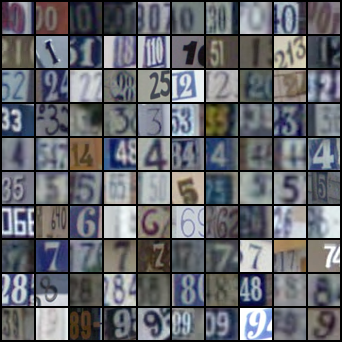}
         \vspace{-15pt}
         \subcaption*{\texttt{original}}
     \end{subfigure}
     \hfill
     \begin{subfigure}{0.19\linewidth}
         \centering
         \includegraphics[width=\linewidth]{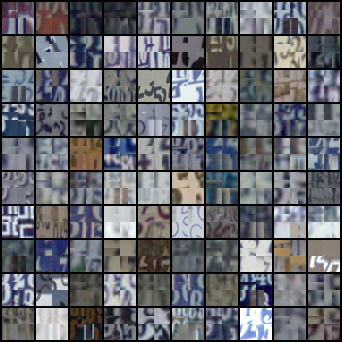}
         \vspace{-15pt}
         \subcaption*{\texttt{jigsaw}}
     \end{subfigure}
     \hfill
     \begin{subfigure}{0.19\linewidth}
         \centering
         \includegraphics[width=\linewidth]{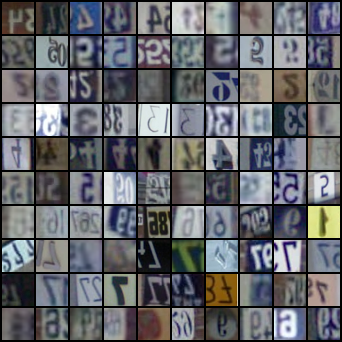}
         \vspace{-15pt}
         \subcaption*{\texttt{flip}}
     \end{subfigure}
     \hfill
     \begin{subfigure}{0.19\linewidth}
         \centering
         \includegraphics[width=\linewidth]{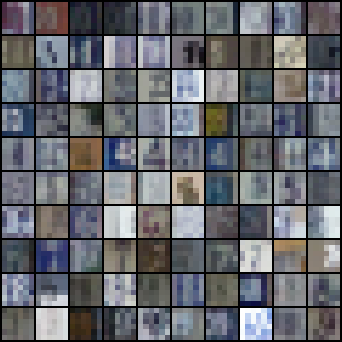}
         \vspace{-15pt}
         \subcaption*{\texttt{mosaic}}
     \end{subfigure}
     \hfill
     \begin{subfigure}{0.19\linewidth}
         \centering
         \includegraphics[width=\linewidth]{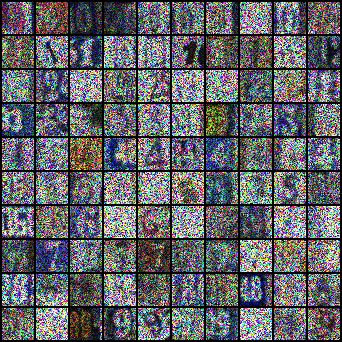}
         \vspace{-15pt}
         \subcaption*{\texttt{speckle}}
     \end{subfigure}
     \vspace{-10pt}
        \caption{Visualization of \emph{pseudo-outliers} on SVHN~\cite{37648}.}
        \label{fig:svhn-poe-visualization}
    \vspace{-5pt}
\end{figure*}

\begin{figure*}[!ht]
     \centering
     \begin{subfigure}{0.19\linewidth}
         \centering
         \includegraphics[width=\linewidth]{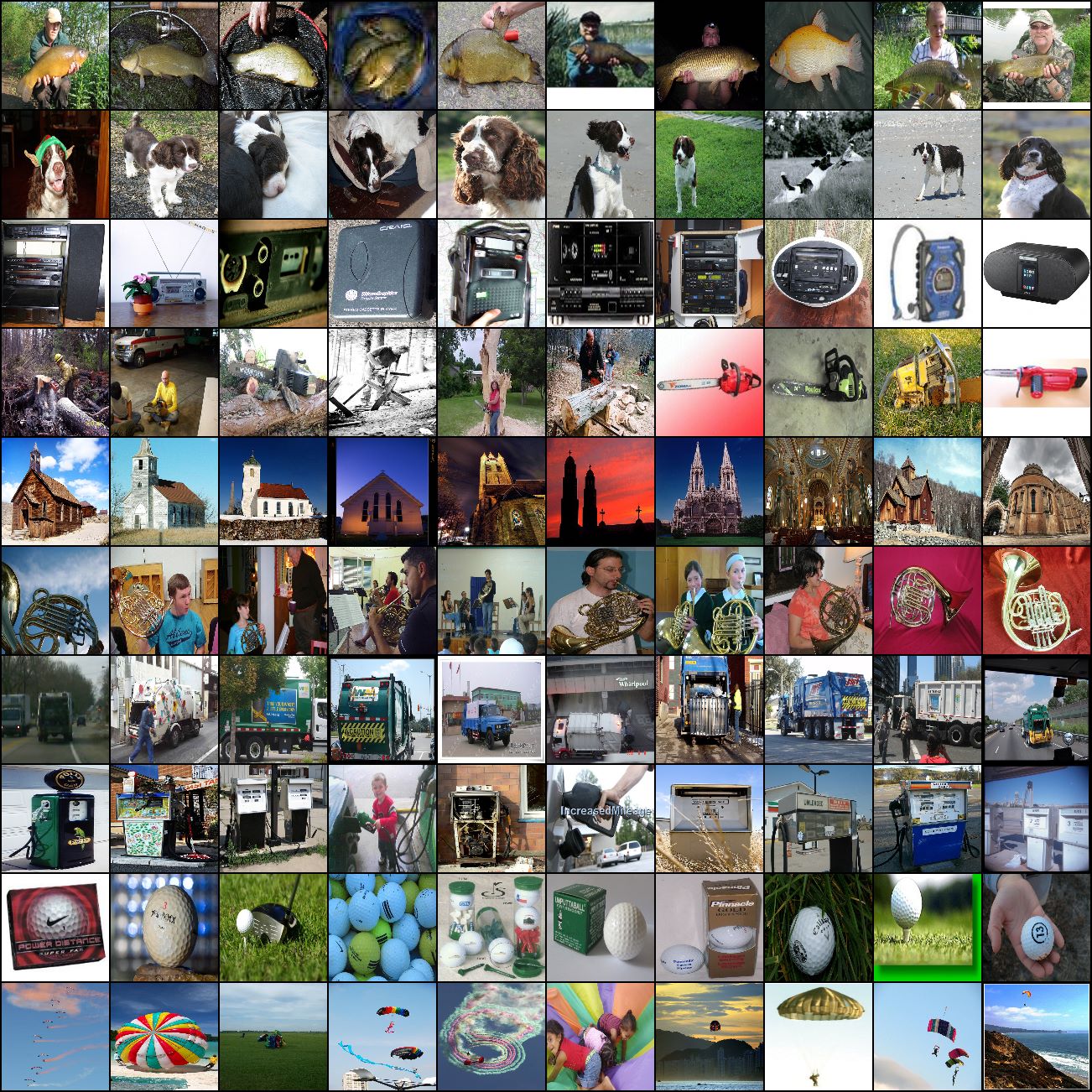}
         \vspace{-15pt}
         \subcaption*{\texttt{original}}
     \end{subfigure}
     \hfill
     \begin{subfigure}{0.19\linewidth}
         \centering
         \includegraphics[width=\linewidth]{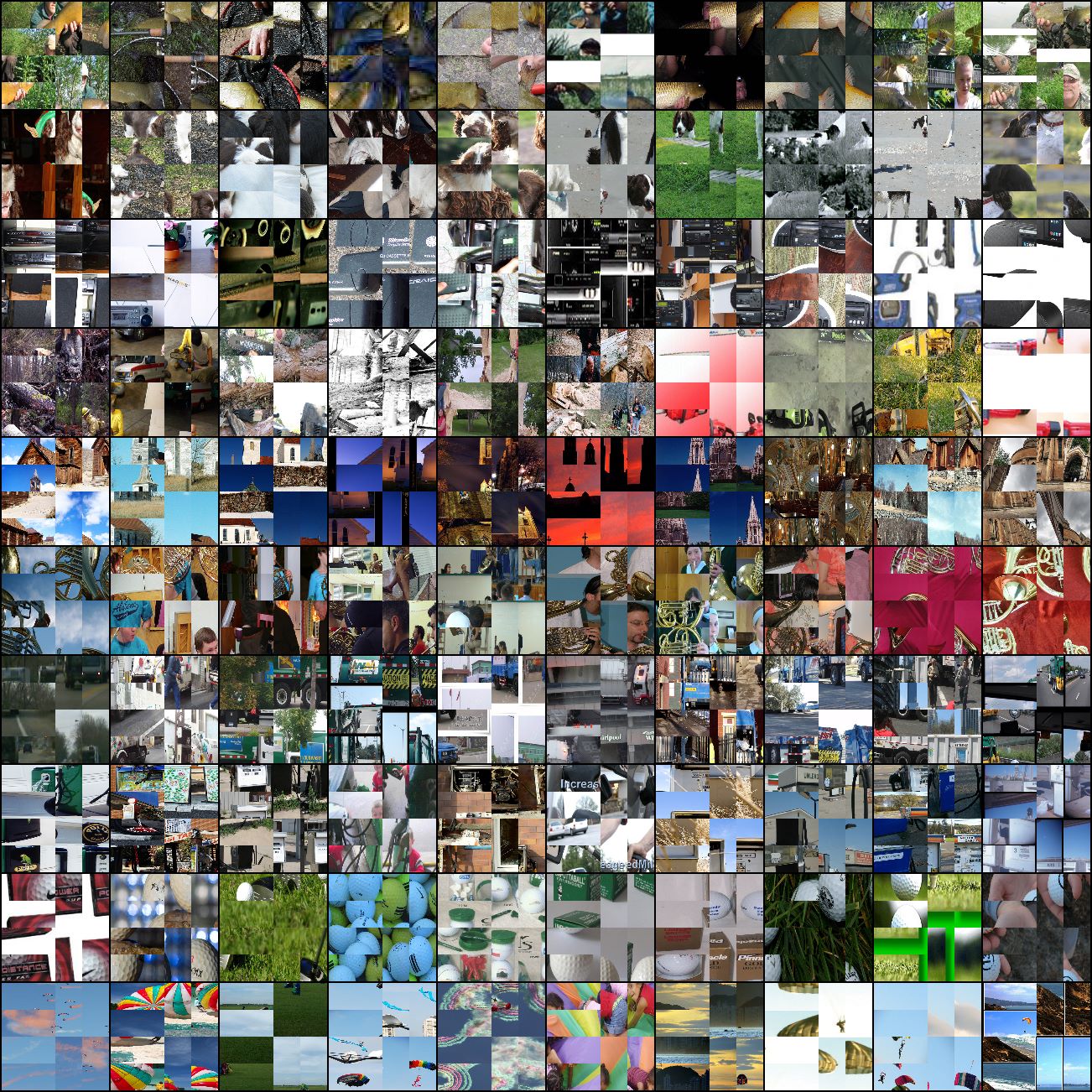}
         \vspace{-15pt}
         \subcaption*{\texttt{jigsaw}}
     \end{subfigure}
     \hfill
     \begin{subfigure}{0.19\linewidth}
         \centering
         \includegraphics[width=\linewidth]{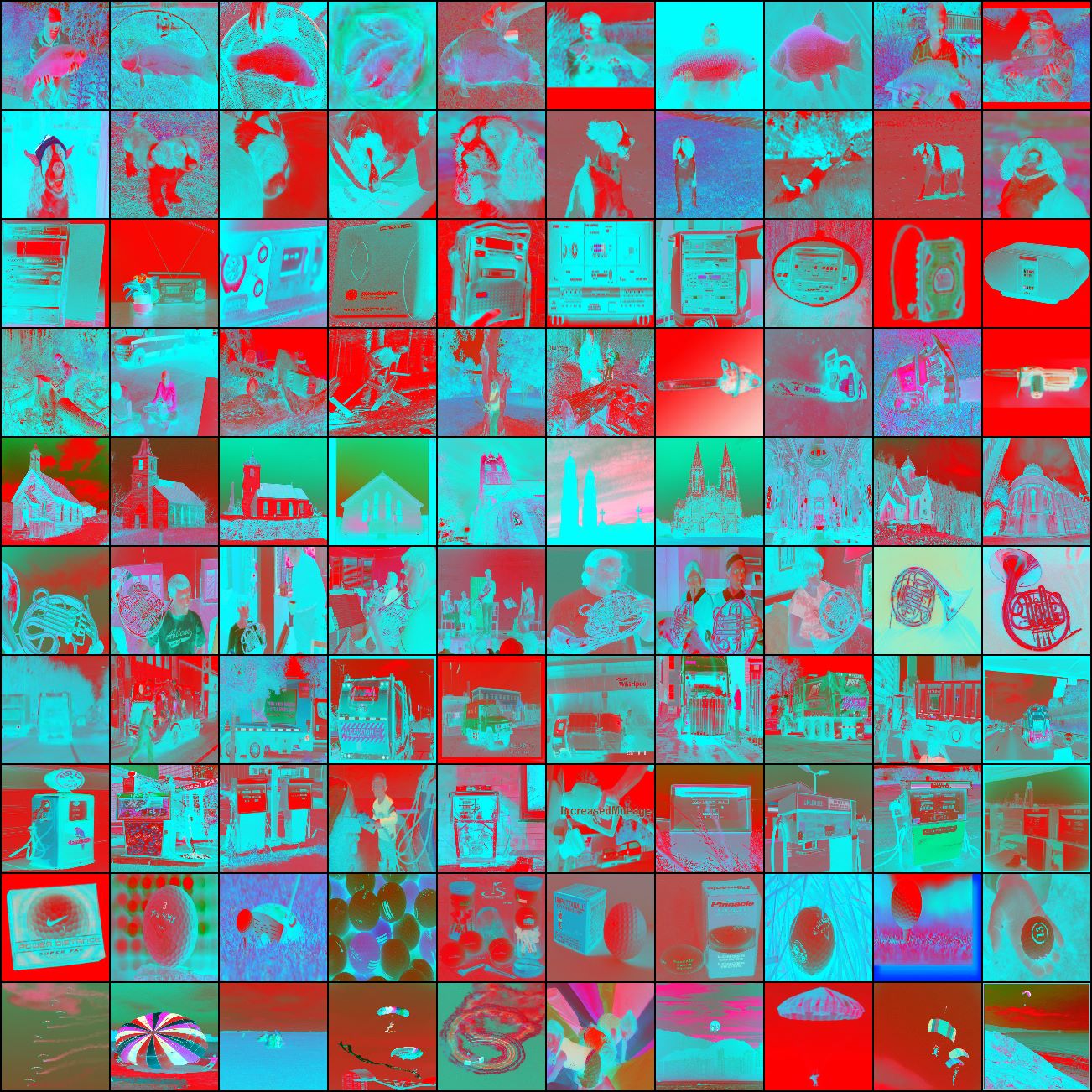}
         \vspace{-15pt}
         \subcaption*{\texttt{invert}}
     \end{subfigure}
     \hfill
     \begin{subfigure}{0.19\linewidth}
         \centering
         \includegraphics[width=\linewidth]{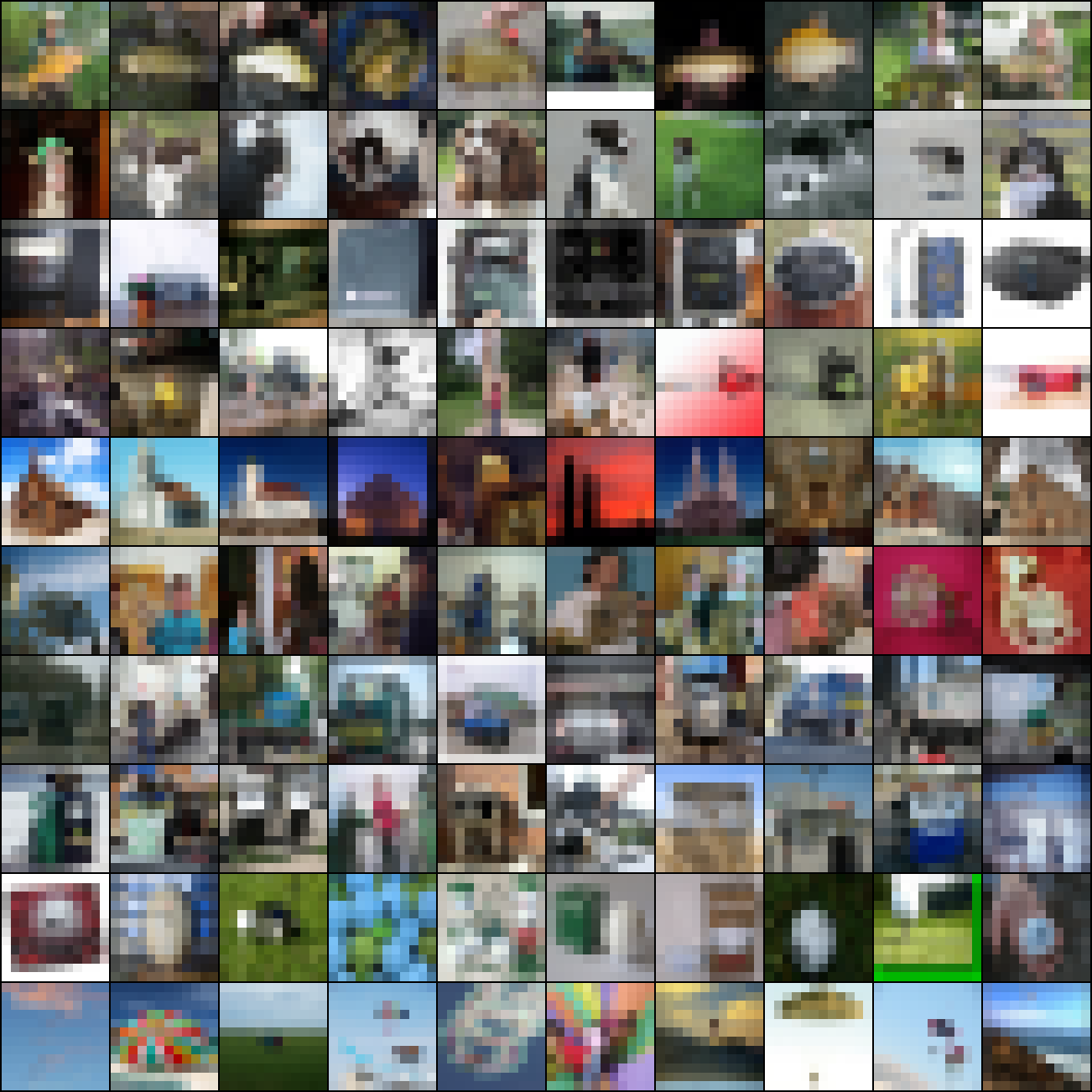}
         \vspace{-15pt}
         \subcaption*{\texttt{mosaic}}
     \end{subfigure}
     \hfill
     \begin{subfigure}{0.19\linewidth}
         \centering
         \includegraphics[width=\linewidth]{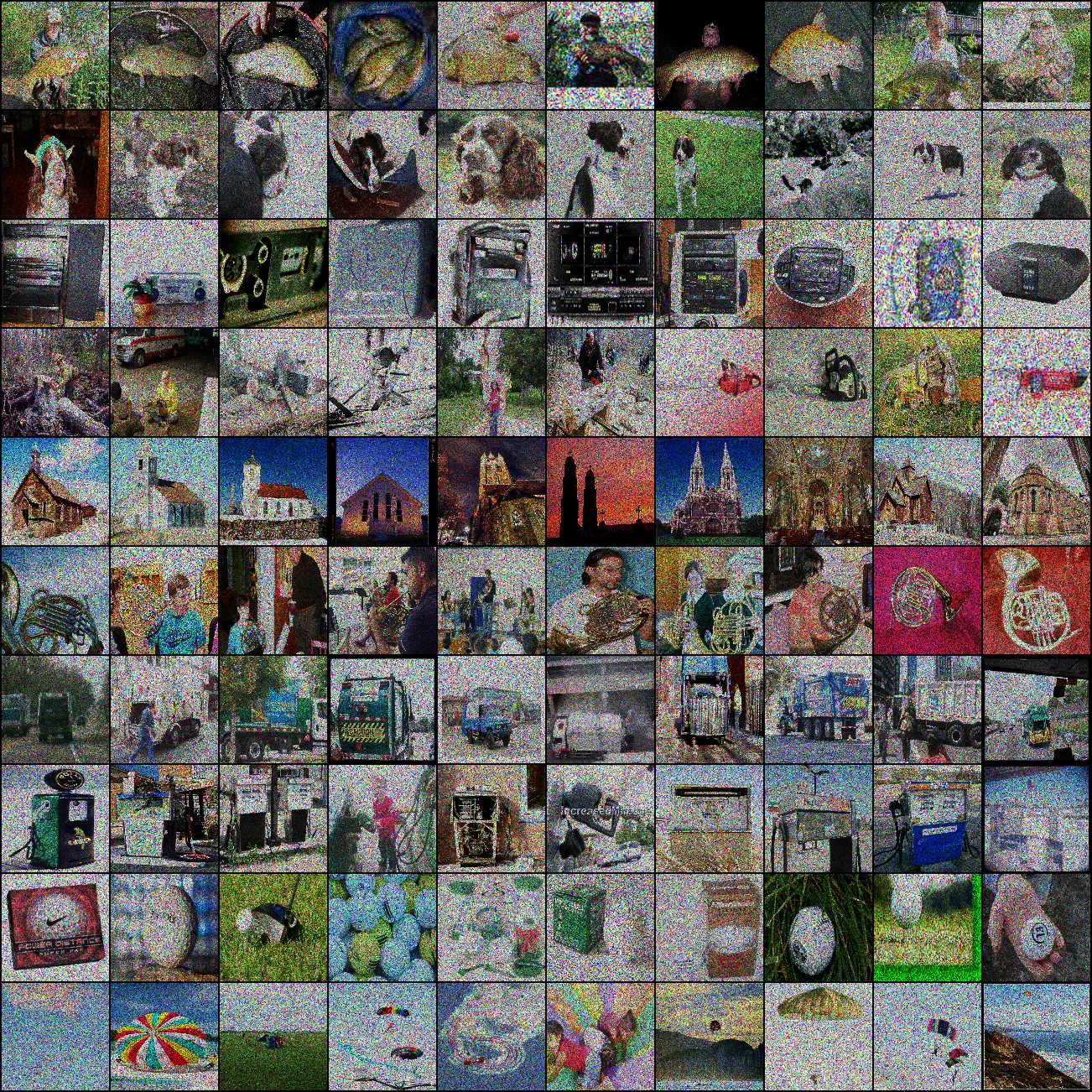}
         \vspace{-15pt}
         \subcaption*{\texttt{speckle}}
     \end{subfigure}
     \vspace{-10pt}
        \caption{Visualization of \emph{pseudo-outliers} on ImageNette~\cite{imagenette}.}
        \label{fig:imagenette-poe-visualization}
    \vspace{-10pt}
\end{figure*}

The \emph{pseudo-outliers} of CIFAR, MNIST, SVHN, and ImageNette are visualized in Figure~\ref{fig:cifar-poe-visualization} --- Figure~\ref{fig:imagenette-poe-visualization}.

\subsection{Performance on Original Full Dataset}
\vspace{-5pt}
We evaluate InD classification and OOD detection on models trained on the original full dataset for reference as theoretical upper bounds, as shown in Table~\ref{tab:full-dataset}. POE still performs competently in the full dataset setting, \ie ordinary OOD detection scenario, which demonstrates the superiority of the proposed POE in enhancing OOD detection, \ie, POE also serves as a competitive method for improving OOD detection performance without the requirement of real outlier dataset. we report the mean OOD performance on the test datasets in Table~\ref{tab:id-ood-list}.

\begin{table*}[!ht]
\setlength\tabcolsep{2pt}
\centering
\renewcommand{\arraystretch}{1}
\caption{InD classification and OOD detection performance of models trained on the original full dataset.}
\label{tab:full-dataset}
\vspace{-7pt}
\resizebox{.9\linewidth}{!}{
\begin{tabular}{@{}cccccccccccccccc@{}}
\toprule
\multirow{2}{*}{Dataset} & \multicolumn{3}{c}{FPR95 $\downarrow$} & \multicolumn{3}{c}{AUROC $\uparrow$} & \multicolumn{3}{c}{AUPR-IN $\uparrow$} & \multicolumn{3}{c}{AUPR-OUT $\uparrow$} & \multicolumn{3}{c}{Accuracy $\uparrow$} \\ \cmidrule(l){2-4} \cmidrule(l){5-7} \cmidrule(l){8-10} \cmidrule(l){11-13} \cmidrule(l){14-16} 
 & baseline & OE & POE (ours) & baseline & OE & POE (ours) & baseline & OE & POE (ours) & baseline & OE & POE (ours) & baseline & OE & POE (ours) \\ \midrule
MNIST & 1.84 & \textbf{0.01} & 0.06 & 99.32 & \textbf{99.99} & 99.92 & 99.47 & \textbf{99.99} & 99.93 & 99.07 & \textbf{99.99} & 99.91 & 99.69 & 99.65 & \textbf{99.74} \\
SVHN & 22.50 & \textbf{0.06} & 2.26 & 96.61 & \textbf{99.98} & 99.56 & 98.67 & \textbf{99.99} & 99.79 & 89.22 & \textbf{99.95} & 98.93 & 95.88 & 95.84 & \textbf{95.90} \\
CIFAR10 & 77.24 & \textbf{35.48} & 35.84 & 79.95 & \textbf{93.21} & 92.25 & 84.08 & \textbf{94.31} & 93.30 & 72.92 & \textbf{91.40} & 90.99 & 85.72 & 86.90 & \textbf{87.10} \\
CIFAR100 & 90.53 & 78.63 & \textbf{60.52} & 66.71 & 77.77 & \textbf{82.09} & 72.72 & 81.87 & \textbf{84.14} & 58.61 & 71.86 & \textbf{79.93} & 57.49 & 58.21 & \textbf{58.64} \\
ImageNette & 77.42 & 73.67 & \textbf{59.06} & 82.91 & 84.52 & \textbf{88.01} & 91.48 & 92.27 & \textbf{93.80} & 65.02 & 67.86 & \textbf{75.57} & 85.80 & 88.00 & \textbf{89.20} \\
ImageFruit & 89.69 & 83.52 & \textbf{79.69} & 67.74 & 69.01 & \textbf{74.87} & 82.03 & 82.32 & \textbf{85.99} & 47.46 & 52.08 & \textbf{59.52} & 64.20 & 65.60 & \textbf{66.40} \\
ImageMisc & 76.88 & 68.59 & \textbf{67.97} & 83.19 & 83.94 & \textbf{84.66} & 91.34 & 91.81 & \textbf{92.36} & 67.09 & 70.11 & \textbf{70.31} & 87.60 & \textbf{88.60} & 87.20 \\ \bottomrule
\end{tabular}
}
\vspace{-5pt}
\end{table*}

\subsection{Additional Results}
\vspace{-5pt}
\subsubsection{Detailed Results of Digit Datasets}
\vspace{-5pt}
Here, we show the detailed results of digit datasets in Table~\ref{tab:digit-full-results}. ConvNets are trained on DSA~\cite{zhao2021DSA} distilled images with IPC=10. The digits contain limited semantics for near-ood \emph{pseudo-outliers} in POE, while OE leverages a real outlier dataset that is rich in semantic information, as a result, OE obtains better OOD performance.

\begin{table*}[!ht]
\setlength\tabcolsep{3pt}
\centering
\renewcommand{\arraystretch}{1}
\caption{OOD detection performance on digit datasets: MNIST~\cite{lecun1998gradient} and SVHN~\cite{37648}, with IPC=10.}
\label{tab:digit-full-results}
\vspace{-7pt}
\resizebox{.9\linewidth}{!}{
\begin{tabular}{@{}cccccccccccccc@{}}
\toprule
\multirow{2}{*}{$\mathcal{D}_{\textrm{in}}$} & \multirow{2}{*}{$\mathcal{D}_\textrm{out}^\textrm{test}$} & \multicolumn{3}{c}{FPR95 $\downarrow$} & \multicolumn{3}{c}{AUROC $\uparrow$} & \multicolumn{3}{c}{AUPR-IN $\uparrow$} & \multicolumn{3}{c}{AUPR-OUT $\uparrow$} \\ \cmidrule(l){3-5} \cmidrule(l){6-8} \cmidrule(l){9-11} \cmidrule(l){12-14} 
 &  & baseline & OE & POE (ours) & baseline & OE & POE (ours) & baseline & OE & POE (ours) & baseline & OE & POE (ours) \\ \midrule
\multirow{7}{*}{MNIST} & Texture & 11.94 & \textbf{0.35} & 1.27 & 97.46 & \textbf{99.94} & 99.01 & 98.81 & \textbf{99.97} & 99.54 & 93.37 & \textbf{99.91} & 97.51 \\
 & Places365 & 20.37 & \textbf{0.06} & 0.28 & 96.80 & \textbf{99.97} & 99.61 & 97.66 & \textbf{99.97} & 99.69 & 95.39 & \textbf{99.97} & 99.50 \\
 & Tiny ImageNet & 21.58 & \textbf{0.08} & 0.36 & 96.65 & \textbf{99.97} & 99.52 & 97.54 & \textbf{99.97} & 99.63 & 95.17 & \textbf{99.97} & 99.38 \\
 & Fashion MNIST & 41.32 & 4.10 & \textbf{1.89} & 92.96 & 99.05 & \textbf{99.22} & 94.42 & 99.19 & \textbf{99.39} & 90.50 & 98.97 & \textbf{99.01} \\
 & notMNIST & 36.36 & \textbf{2.67} & 6.59 & 93.75 & \textbf{99.33} & 98.58 & 94.72 & \textbf{99.40} & 98.81 & 92.00 & \textbf{99.31} & 98.30 \\
 & CIFAR10 & 22.69 & \textbf{0.00} & 0.31 & 96.56 & \textbf{99.99} & 99.56 & 97.49 & \textbf{99.99} & 99.66 & 94.98 & \textbf{99.99} & 99.42 \\ \cmidrule(l){2-14} 
 & \textbf{mean} & 25.71 & \textbf{1.21} & 1.78 & 95.70 & \textbf{99.71} & 99.25 & 96.77 & \textbf{99.75} & 99.45 & 93.57 & \textbf{99.69} & 98.85 \\ \midrule
\multirow{7}{*}{SVHN} & Texture & 69.73 & \textbf{23.07} & 29.77 & 83.66 & \textbf{95.95} & 94.81 & 96.09 & \textbf{99.08} & 98.83 & 49.87 & \textbf{85.20} & 80.81 \\
 & Places365 & 71.93 & \textbf{23.73} & 36.73 & 83.63 & \textbf{96.01} & 93.72 & 86.69 & \textbf{96.57} & 94.71 & 79.92 & \textbf{95.48} & 92.63 \\
 & Tiny ImageNet & 70.29 & \textbf{16.24} & 18.90 & 84.39 & \textbf{97.27} & 96.89 & 93.97 & \textbf{98.98} & 98.84 & 62.98 & \textbf{92.88} & 91.96 \\
 & LSUN & 68.88 & \textbf{11.09} & 13.95 & 84.93 & \textbf{98.04} & 97.62 & 94.18 & \textbf{99.27} & 99.11 & 64.08 & \textbf{94.79} & 93.80 \\
 & iSUN & 69.32 & \textbf{11.02} & 14.03 & 84.83 & \textbf{98.04} & 97.61 & 94.70 & \textbf{99.35} & 99.20 & 61.25 & \textbf{94.26} & 93.14 \\
 & CIFAR10 & 72.82 & \textbf{27.80} & 42.00 & 83.34 & \textbf{95.19} & 92.48 & 93.53 & \textbf{98.16} & 97.12 & 60.78 & \textbf{88.18} & 81.34 \\ \cmidrule(l){2-14} 
 & \textbf{mean} & 70.50 & \textbf{18.83} & 25.90 & 84.13 & \textbf{96.75} & 95.52 & 93.19 & \textbf{98.57} & 97.97 & 63.15 & \textbf{91.80} & 88.95 \\ \bottomrule
\end{tabular}
}
\vspace{-5pt}
\end{table*}

\begin{table*}[!ht]
\setlength\tabcolsep{3pt}
\centering
\renewcommand{\arraystretch}{1}
\caption{OOD detection performance on ImageNet Subsets~\cite{cazenavette2022distillation,imagenette}, with IPC=1.}
\label{tab:imagenet-full-results}
\vspace{-7pt}
\resizebox{.9\linewidth}{!}{
\begin{tabular}{@{}cccccccccccccc@{}}
\toprule
\multirow{2}{*}{$\mathcal{D}_{\textrm{in}}$} & \multirow{2}{*}{$\mathcal{D}_\textrm{out}^\textrm{test}$} & \multicolumn{3}{c}{FPR95 $\downarrow$} & \multicolumn{3}{c}{AUROC $\uparrow$} & \multicolumn{3}{c}{AUPR-IN $\uparrow$} & \multicolumn{3}{c}{AUPR-OUT $\uparrow$} \\ \cmidrule(l){3-5} \cmidrule(l){6-8} \cmidrule(l){9-11} \cmidrule(l){12-14} 
 &  & baseline & OE & POE(ours) & baseline & OE & POE(ours) & baseline & OE & POE(ours) & baseline & OE & POE(ours) \\ \midrule
\multirow{6}{*}{ImageNette} & Texture & 90.31 & 84.53 & \textbf{80.47} & 63.80 & 69.65 & \textbf{72.98} & 76.98 & 82.42 & \textbf{83.95} & 45.56 & 51.32 & \textbf{56.35} \\
 & Species & 93.59 & 91.48 & \textbf{89.53} & 56.62 & 64.22 & \textbf{65.21} & 71.06 & \textbf{79.05} & 79.02 & 39.24 & 43.86 & \textbf{45.84} \\
 & iNaturalist & 92.34 & 88.05 & \textbf{87.73} & 59.96 & 64.87 & \textbf{68.19} & 73.59 & 78.67 & \textbf{81.03} & 41.59 & 45.98 & \textbf{48.81} \\
 & ImageNet-O & 93.20 & 91.17 & \textbf{89.06} & 59.13 & \textbf{63.69} & 62.68 & 74.07 & \textbf{78.67} & 76.56 & 40.14 & 43.83 & \textbf{44.58} \\
 & OpenImage-O & 93.83 & 89.92 & \textbf{86.33} & 58.09 & 64.34 & \textbf{67.11} & 72.71 & 78.65 & \textbf{79.90} & 39.68 & 44.86 & \textbf{49.53} \\ \cmidrule(l){2-14} 
 & \textbf{mean} & 92.66 & 89.03 & \textbf{86.62} & 59.52 & 65.35 & \textbf{67.23} & 73.68 & 79.49 & \textbf{80.09} & 41.24 & 45.97 & \textbf{49.02} \\ \midrule
\multirow{6}{*}{ImageFruit} & Texture & 92.73 & 90.70 & \textbf{90.16} & 54.48 & 58.97 & \textbf{61.71} & 69.99 & 73.53 & \textbf{75.77} & 38.10 & 41.93 & \textbf{44.61} \\
 & Species & 95.70 & \textbf{91.48} & 92.58 & 51.39 & \textbf{58.91} & 56.12 & 67.99 & \textbf{73.80} & 71.45 & 34.93 & \textbf{41.02} & 39.56 \\
 & iNaturalist & 94.53 & 91.33 & \textbf{91.17} & 52.53 & 55.90 & \textbf{60.30} & 68.37 & 71.57 & \textbf{75.10} & 36.21 & 38.99 & \textbf{42.38} \\
 & ImageNet-O & 95.31 & 92.66 & \textbf{92.27} & 49.13 & \textbf{54.74} & 53.07 & 65.57 & \textbf{69.91} & 68.37 & 34.06 & 37.72 & \textbf{37.93} \\
 & OpenImage-O & 94.38 & \textbf{92.73} & 93.91 & 48.37 & \textbf{55.99} & 54.50 & 64.94 & \textbf{71.45} & 70.35 & 33.93 & \textbf{38.88} & 37.18 \\ \cmidrule(l){2-14} 
 & \textbf{mean} & 94.53 & \textbf{91.78} & 92.02 & 51.18 & 56.90 & \textbf{57.14} & 67.37 & 72.05 & \textbf{72.21} & 35.45 & 39.71 & \textbf{40.33} \\ \midrule
\multirow{6}{*}{ImageMisc} & Texture & 90.86 & 86.09 & \textbf{85.55} & 62.33 & 68.04 & \textbf{69.36} & 77.68 & 81.90 & \textbf{82.62} & 43.30 & 49.73 & \textbf{51.86} \\
 & Species & 93.59 & 92.97 & \textbf{90.47} & 57.99 & \textbf{62.85} & 62.67 & 74.98 & \textbf{78.89} & 77.77 & 38.99 & 42.45 & \textbf{43.58} \\
 & iNaturalist & 93.36 & 89.61 & \textbf{88.75} & 59.17 & 66.64 & \textbf{68.54} & 76.31 & 80.73 & \textbf{81.76} & 39.56 & 47.02 & \textbf{49.25} \\
 & ImageNet-O & 94.14 & \textbf{93.20} & 93.36 & 56.66 & \textbf{61.50} & 58.81 & 73.69 & \textbf{77.07} & 75.34 & 38.13 & \textbf{41.46} & 40.12 \\
 & OpenImage-O & 93.28 & \textbf{90.39} & 91.56 & 58.28 & \textbf{63.19} & 60.65 & 74.93 & \textbf{78.18} & 75.77 & 39.43 & \textbf{44.55} & 41.87 \\ \cmidrule(l){2-14} 
 & \textbf{mean} & 93.05 & 90.45 & \textbf{89.94} & 58.88 & \textbf{64.44} & 64.01 & 75.52 & \textbf{79.36} & 78.66 & 39.88 & 45.04 & \textbf{45.33} \\ \bottomrule
\end{tabular}
}
\vspace{-5pt}
\end{table*}

\subsubsection{Detailed Results of ImageNet Subsets}
Results on three ImageNet Subsets are shown in Table~\ref{tab:imagenet-full-results}. ConvNets are trained on DSA~\cite{zhao2021DSA} distilled images with IPC=1. POE consistently outperforms OE in most circumstances, which illustrates the effectiveness of POE.

\clearpage
\subsection{Additional Analysis}

\subsubsection{Corruptions Transformations}

In this paper, for POE, we mainly implement several corruptions and utilize the ensemble of \emph{pseudo-outliers} from various corruption operations. We also test the performance of POE with only one certain type of corruption, as shown in Table~\ref{tab:corruption-type}. The experimental results demonstrate the benefits of the ensemble of these corruption transformations. The reason is that the ensemble offers more near-ood samples for better modelling the boundaries between InD and OOD samples, leading to better OOD performance.

\begin{table}[htbp]
\setlength\tabcolsep{3pt}
\centering
\renewcommand{\arraystretch}{1}
\caption{OOD detection performance of POE with different corruption transformations.}
\label{tab:corruption-type}
\vspace{-7pt}
\resizebox{\linewidth}{!}{
\begin{tabular}{@{}ccccccc@{}}
\toprule
corruption & \texttt{no} & \texttt{jigsaw} & \texttt{invert} & \texttt{mosaic} & \texttt{speckle} & \texttt{ensemble} \\ \midrule
AUROC      & 64.06       & 69.33           & 66.48           & 64.74           & 70.67            & \textbf{72.50}    \\ \bottomrule
\end{tabular}
}
\end{table}

\subsubsection{Why not distill both InD and outliers into InD?}

Here, let's continue the third issue in Section~\ref{subsec:further-analysis}. One seemingly more efficient way is to distill $\mathcal{T}_\textrm{in}$ and $\mathcal{T}_\textrm{out}$ altogether into a single set $\mathcal{S}_\textrm{single}$ rather than into $\mathcal{S}_\textrm{in}$ and $\mathcal{S}_\textrm{out}$ separately, to be more specific, one could simply leave out the second term in Eq.~\eqref{eq:loss-on-distilled-set}, \ie, make model trained on $\mathcal{S}_\textrm{single}$ could approximate the performance when trained on both $\mathcal{T}_\textrm{in}$ and $\mathcal{T}_\textrm{out}$. For example, when implementing on DSA~\cite{zhao2021DSA}, one could rewrite the distillation loss as:
\begin{equation}
    \mathcal{L}_\textrm{distill}(\mathcal{S})=D\big(\nabla_\theta\mathcal{L}_\textrm{ce}(\mathcal{A}(\mathcal{S}_\textrm{single}),\hat\theta_t),\nabla_\theta\mathcal{L}(\mathcal{A}(\mathcal{T}),\hat\theta_t)\big)
    \label{eq:distill-loss-single}
\end{equation}
where $\mathcal{L}$ is the \emph{integrated loss} considering both tasks: $\mathcal{L}=\mathbb{E}_{(\boldsymbol{x},y)\sim \mathcal{D}_\textrm{in}}\mathcal{L}_\textrm{ce}(f_\theta(\boldsymbol{x}),y)+\lambda\mathbb{E}_{\boldsymbol{x}^\prime\sim\mathcal{D}_\textrm{out}}H(\mathcal{U};f_\theta(\boldsymbol{x}^\prime))$, and $\mathcal{L}_\textrm{ce}$ is the cross-entropy loss. Eq.~\eqref{eq:distill-loss-single} aims to make the gradient trained on $\mathcal{S}_\textrm{single}$ approaches the gradient trained on both $\mathcal{T}_\textrm{in}$ and $\mathcal{T}_\textrm{out}$, which is more efficient.

\begin{figure}
    \centering
    \begin{subfigure}{.48\linewidth}
        \includegraphics[width=\linewidth]{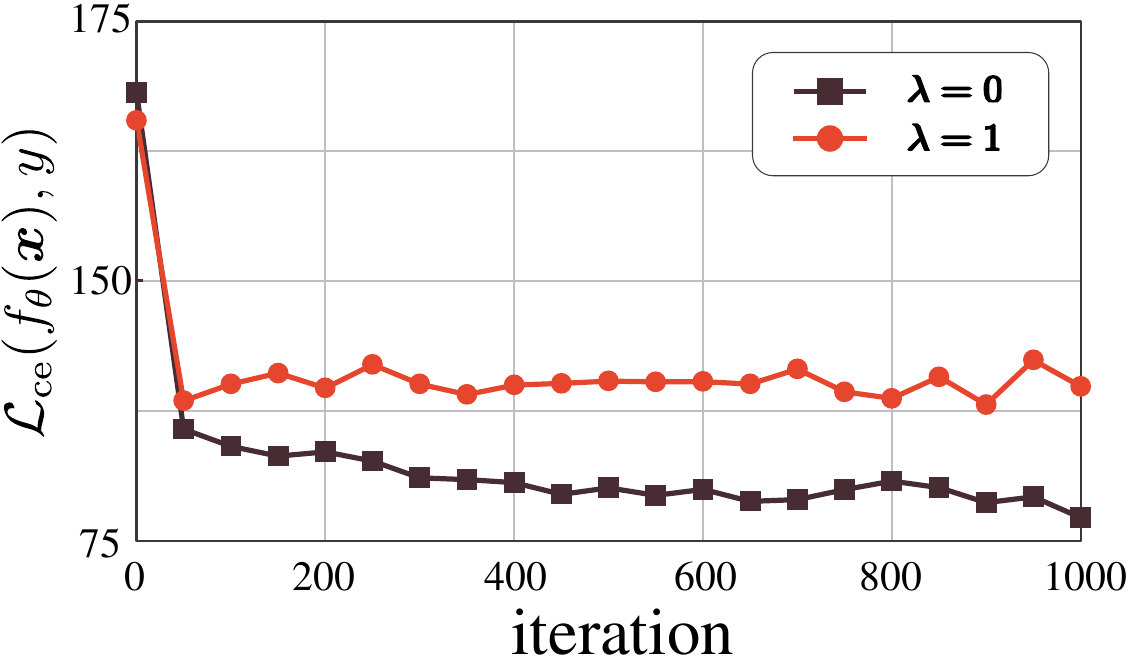}
        \vspace{-15pt}
    \caption{InD loss $\mathcal{L}_\textrm{ce}(f_\theta(\boldsymbol{x}),y)$}
        \label{subfig:ce-loss-iter}
    \end{subfigure}
    \hfill
    \begin{subfigure}{.48\linewidth}
        \includegraphics[width=\linewidth]{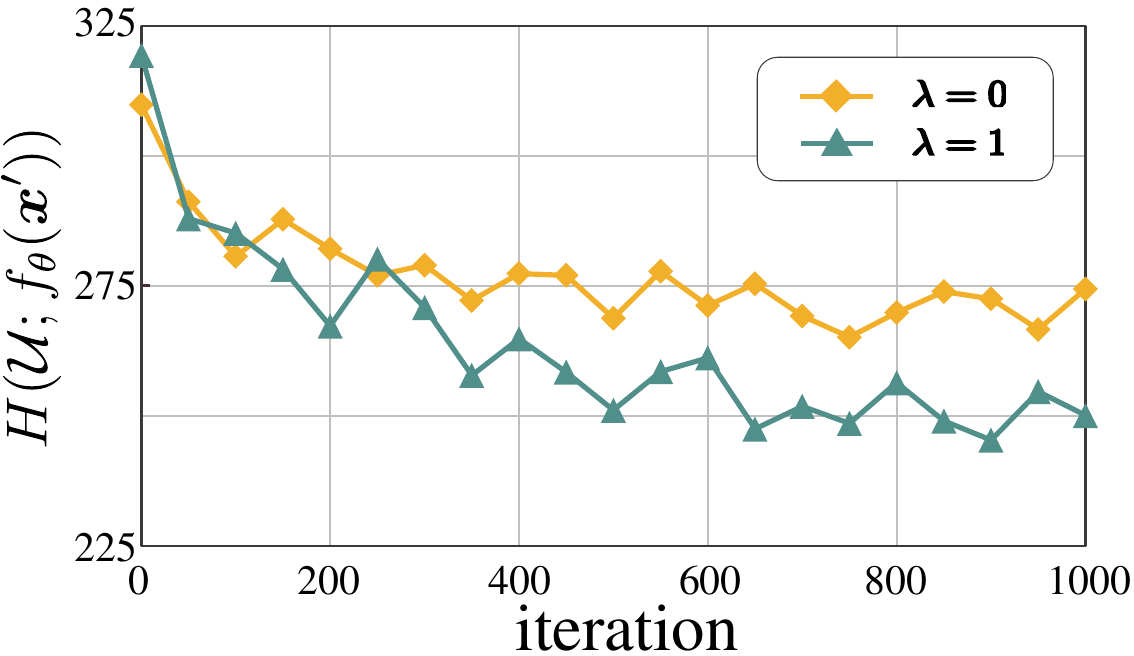}
        \vspace{-15pt}
        \caption{OOD loss $H(\mathcal{U};f_\theta(\boldsymbol{x}^\prime))$}
        \label{subfig:h-loss-iter}
    \end{subfigure}
    \vspace{-10pt}
    \caption{InD and OOD loss function with respect to different iterations during training. $\lambda=0$ denotes baseline, and $\lambda=1$ means that distill both InD and OOD into a single set $\mathcal{S}_\textrm{single}$.}
    \label{fig:loss-iter}
    \vspace{-10pt}
\end{figure}

However, this does not work as expected. From Figure~\ref{fig:loss-iter} one could tell that when distilling  $\mathcal{T}_\textrm{in}$ and $\mathcal{T}_\textrm{out}$ jointly to a single set $\mathcal{S}_\textrm{single}$, the classification loss $\mathcal{L}_\textrm{ce}$ increases compared with baseline (\ie $\lambda=0$), although the outlier loss $H(\cdot)$ decreases, in this case, the InD classification drops from $\sim52\%$ to $\sim40\%$, leading to drastically weak OOD performance (AUROC dropped from $\sim 64\%$ to $60\%$). An intuitive explanation is that, as in Section~\ref{subsec:further-analysis}, the outlier loss $H(\cdot)$ decreased compared with the baseline at the cost of the increase in classification loss $\mathcal{L}_\textrm{ce}$, \ie, at the cost of InD classification performance, which could further deteriorate OOD performance due to the approximately positive correlation~\cite{vaze2022openset} between InD and OOD performance. When keeping the total distilled size $|\mathcal{S}|$ fixed, additionally incorporating outlier information into $\mathcal{S}_\textrm{single}$ will lead to a decrease in the information content of in-distribution portion in $\mathcal{S}_\textrm{single}$.

\newpage
\subsection{Additional Visualizations} \label{subsec:appendix-visualize}
\begin{figure*}[!ht]
     \centering
     \begin{subfigure}{0.45\linewidth}
         \centering
         \includegraphics[width=\linewidth]{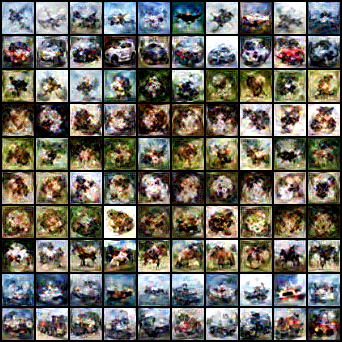}
         \subcaption*{OE: $\mathcal{S}_\textrm{in}$}
     \end{subfigure}
     \hfill
     \begin{subfigure}{0.45\linewidth}
         \centering
         \includegraphics[width=\linewidth]{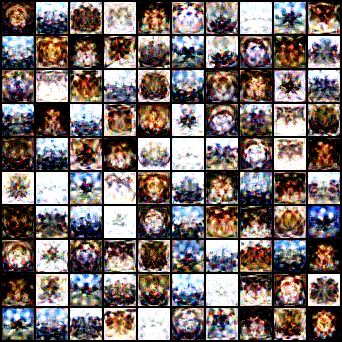}
         \subcaption*{OE: $\mathcal{S}_\textrm{out}$}
     \end{subfigure}
     \vskip\baselineskip
     \begin{subfigure}{0.45\linewidth}
         \centering
         \includegraphics[width=\linewidth]{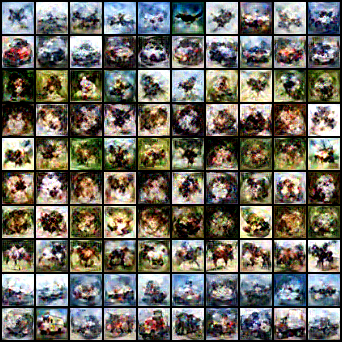}
         \subcaption*{POE: $\mathcal{S}_\textrm{in}$}
     \end{subfigure}
     \hfill
     \begin{subfigure}{0.45\linewidth}
         \centering
         \includegraphics[width=\linewidth]{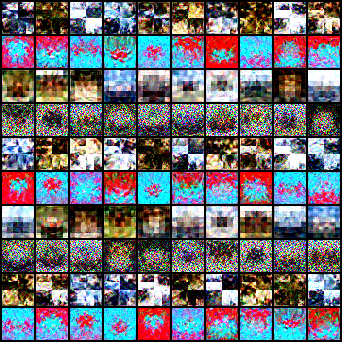}
         \subcaption*{POE: $\mathcal{S}_\textrm{out}$}
     \end{subfigure}
        \caption{Visualization of $\mathcal{S}_\textrm{in}$ and $\mathcal{S}_\textrm{out}$ distilled by TrustDD with OE and POE on CIFAR10~\cite{krizhevsky2009learning}, with IPC=10.}
        \label{fig:cifar-more-visualization}
    \vspace{-5pt}
\end{figure*}

\begin{figure*}[!ht]
     \centering
     \begin{subfigure}{0.45\linewidth}
         \centering
         \includegraphics[width=\linewidth]{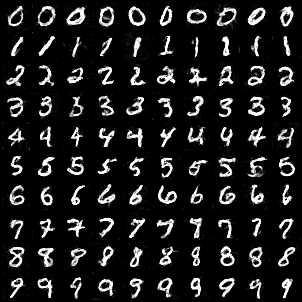}
         \subcaption*{OE: $\mathcal{S}_\textrm{in}$}
     \end{subfigure}
     \hfill
     \begin{subfigure}{0.45\linewidth}
         \centering
         \includegraphics[width=\linewidth]{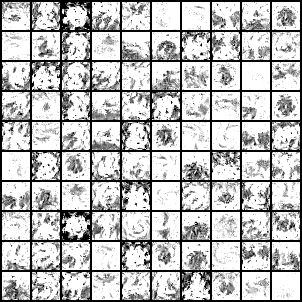}
         \subcaption*{OE: $\mathcal{S}_\textrm{out}$}
     \end{subfigure}
     \vskip\baselineskip
     \begin{subfigure}{0.45\linewidth}
         \centering
         \includegraphics[width=\linewidth]{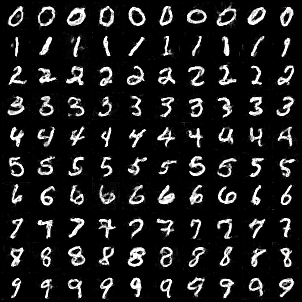}
         \subcaption*{POE: $\mathcal{S}_\textrm{in}$}
     \end{subfigure}
     \hfill
     \begin{subfigure}{0.45\linewidth}
         \centering
         \includegraphics[width=\linewidth]{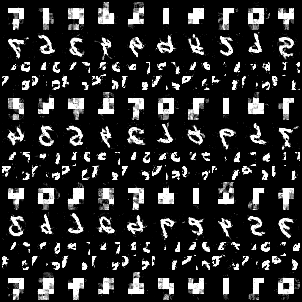}
         \subcaption*{POE: $\mathcal{S}_\textrm{out}$}
     \end{subfigure}
        \caption{Visualization of $\mathcal{S}_\textrm{in}$ and $\mathcal{S}_\textrm{out}$ distilled by TrustDD with OE and POE on MNIST~\cite{lecun1998gradient}, with IPC=10.}
        \label{fig:mnist-more-visualization}
    \vspace{-5pt}
\end{figure*}

\begin{figure*}[!ht]
     \centering
     \begin{subfigure}{0.45\linewidth}
         \centering
         \includegraphics[width=\linewidth]{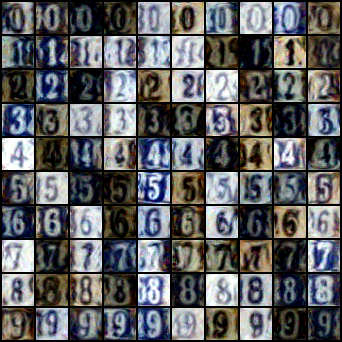}
         \subcaption*{OE: $\mathcal{S}_\textrm{in}$}
     \end{subfigure}
     \hfill
     \begin{subfigure}{0.45\linewidth}
         \centering
         \includegraphics[width=\linewidth]{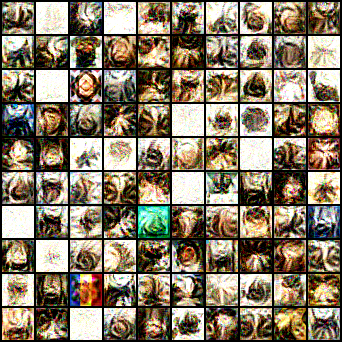}
         \subcaption*{OE: $\mathcal{S}_\textrm{out}$}
     \end{subfigure}
     \vskip\baselineskip
     \begin{subfigure}{0.45\linewidth}
         \centering
         \includegraphics[width=\linewidth]{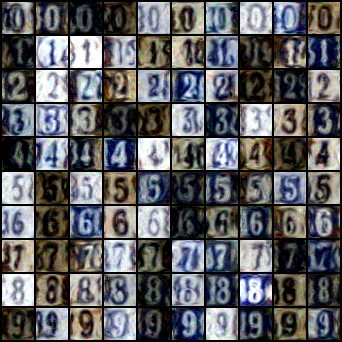}
         \subcaption*{POE: $\mathcal{S}_\textrm{in}$}
     \end{subfigure}
     \hfill
     \begin{subfigure}{0.45\linewidth}
         \centering
         \includegraphics[width=\linewidth]{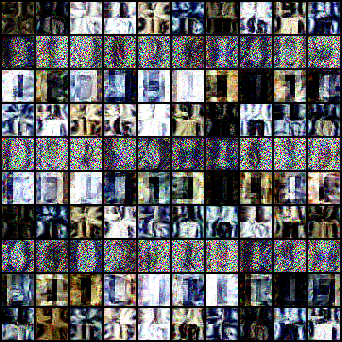}
         \subcaption*{POE: $\mathcal{S}_\textrm{out}$}
     \end{subfigure}
     \vspace{-10pt}
        \caption{Visualization of $\mathcal{S}_\textrm{in}$ and $\mathcal{S}_\textrm{out}$ distilled by TrustDD with OE and POE on SVHN~\cite{37648}, with IPC=10.}
        \label{fig:svhn-more-visualization}
    \vspace{-5pt}
\end{figure*}

\begin{figure*}[!ht]
     \centering
     \begin{subfigure}{0.45\linewidth}
         \centering
         \includegraphics[width=\linewidth]{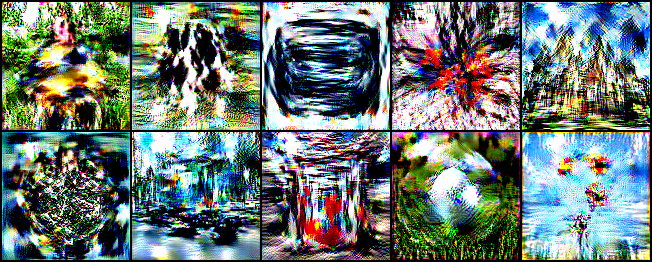}
         \subcaption*{OE: $\mathcal{S}_\textrm{in}$}
     \end{subfigure}
     \hfill
     \begin{subfigure}{0.45\linewidth}
         \centering
         \includegraphics[width=\linewidth]{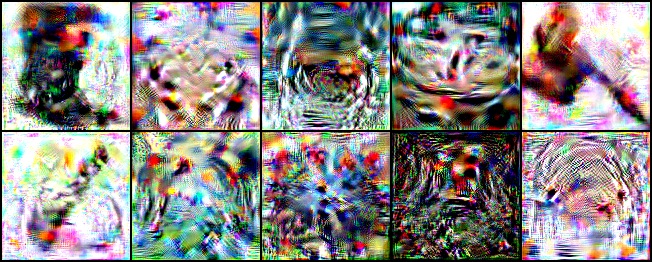}
         \subcaption*{OE: $\mathcal{S}_\textrm{out}$}
     \end{subfigure}
     \vskip\baselineskip
     \begin{subfigure}{0.45\linewidth}
         \centering
         \includegraphics[width=\linewidth]{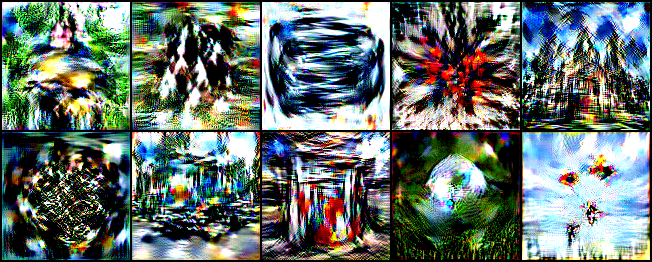}
         \subcaption*{POE: $\mathcal{S}_\textrm{in}$}
     \end{subfigure}
     \hfill
     \begin{subfigure}{0.45\linewidth}
         \centering
         \includegraphics[width=\linewidth]{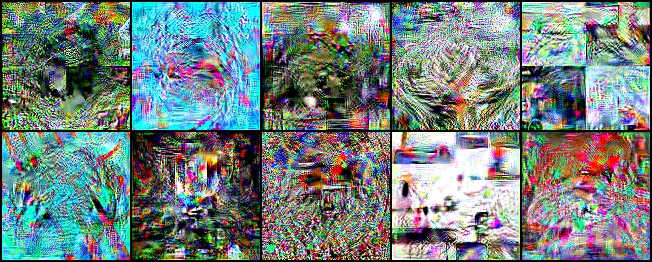}
         \subcaption*{POE: $\mathcal{S}_\textrm{out}$}
     \end{subfigure}
        \caption{Visualization of $\mathcal{S}_\textrm{in}$ and $\mathcal{S}_\textrm{out}$ distilled by TrustDD with OE and POE on ImageNette~\cite{imagenette}, with IPC=1.}
        \label{fig:imagenette-more-visualization}
    \vspace{-5pt}
\end{figure*}

\begin{figure*}[!ht]
     \centering
     \begin{subfigure}{0.45\linewidth}
         \centering
         \includegraphics[width=\linewidth]{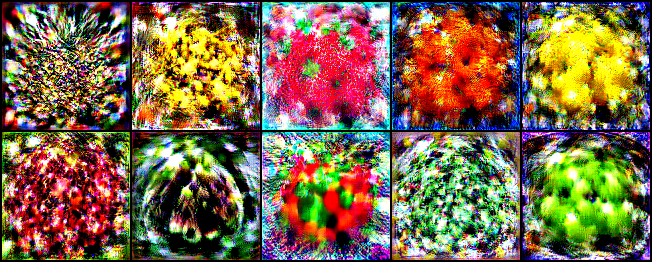}
         \subcaption*{OE: $\mathcal{S}_\textrm{in}$}
     \end{subfigure}
     \hfill
     \begin{subfigure}{0.45\linewidth}
         \centering
         \includegraphics[width=\linewidth]{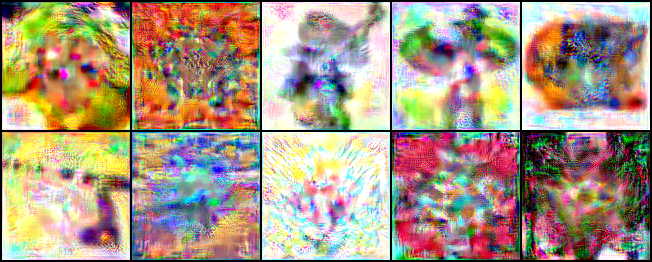}
         \subcaption*{OE: $\mathcal{S}_\textrm{out}$}
     \end{subfigure}
     \vskip\baselineskip
     \begin{subfigure}{0.45\linewidth}
         \centering
         \includegraphics[width=\linewidth]{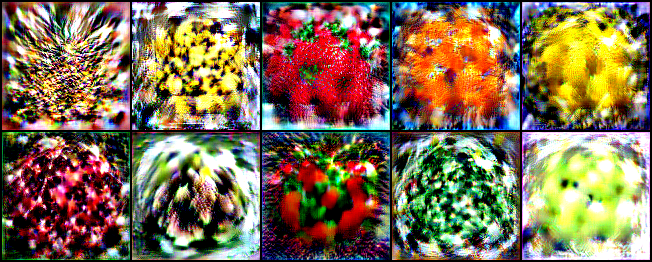}
         \subcaption*{POE: $\mathcal{S}_\textrm{in}$}
     \end{subfigure}
     \hfill
     \begin{subfigure}{0.45\linewidth}
         \centering
         \includegraphics[width=\linewidth]{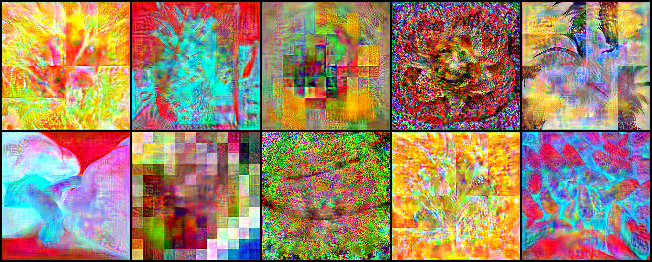}
         \subcaption*{POE: $\mathcal{S}_\textrm{out}$}
     \end{subfigure}
        \caption{Visualization of $\mathcal{S}_\textrm{in}$ and $\mathcal{S}_\textrm{out}$ distilled by TrustDD with OE and POE on ImageFruit~\cite{cazenavette2022distillation}, with IPC=1.}
        \label{fig:imagefruit-more-visualization}
    \vspace{-5pt}
\end{figure*}

\begin{figure*}[!ht]
     \centering
     \begin{subfigure}{0.45\linewidth}
         \centering
         \includegraphics[width=\linewidth]{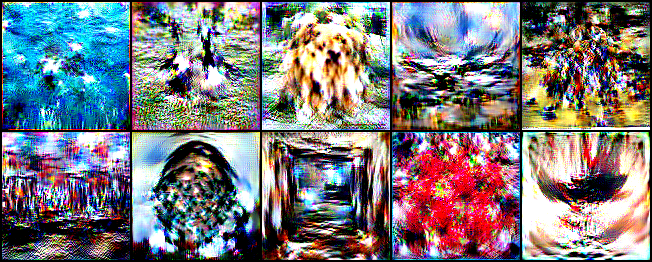}
         \subcaption*{OE: $\mathcal{S}_\textrm{in}$}
     \end{subfigure}
     \hfill
     \begin{subfigure}{0.45\linewidth}
         \centering
         \includegraphics[width=\linewidth]{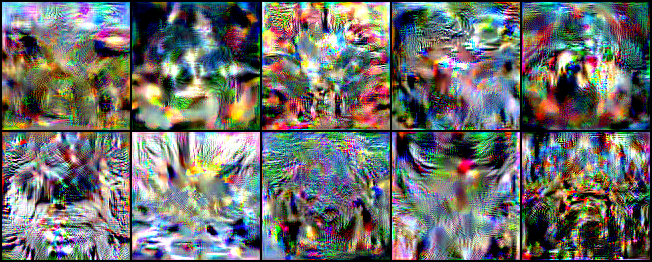}
         \subcaption*{OE: $\mathcal{S}_\textrm{out}$}
     \end{subfigure}
     \vskip\baselineskip
     \begin{subfigure}{0.45\linewidth}
         \centering
         \includegraphics[width=\linewidth]{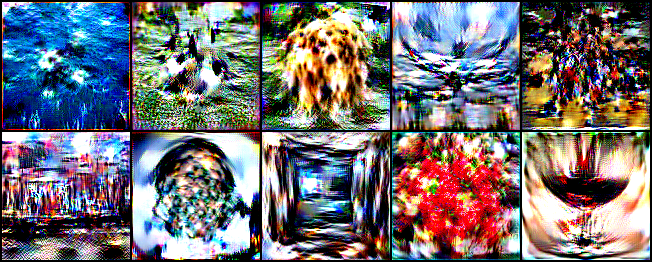}
         \subcaption*{POE: $\mathcal{S}_\textrm{in}$}
     \end{subfigure}
     \hfill
     \begin{subfigure}{0.45\linewidth}
         \centering
         \includegraphics[width=\linewidth]{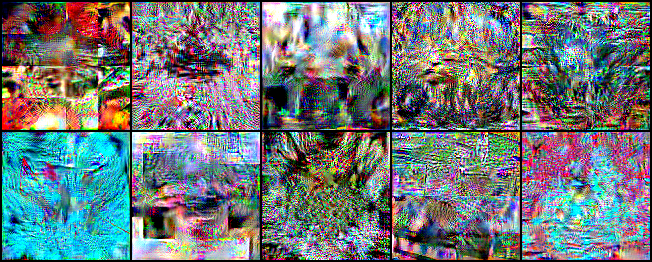}
         \subcaption*{POE: $\mathcal{S}_\textrm{out}$}
     \end{subfigure}
        \caption{Visualization of $\mathcal{S}_\textrm{in}$ and $\mathcal{S}_\textrm{out}$ distilled by TrustDD with OE and POE on ImageMisc, with IPC=1.}
        \label{fig:imagemisc-more-visualization}
    \vspace{-5pt}
\end{figure*}

In this section, we visualize additional results to complement the results of Section~\ref{subsec:visualization}. Visualizations of CIFAR10, MNIST, SVHN, ImageNette, ImageFruit and ImageMisc are shown in Figure~\ref{fig:cifar-more-visualization}~---~Figure~\ref{fig:imagemisc-more-visualization}.

\clearpage

\end{document}